\documentclass{kyrielei}
\usepackage{amsmath}

\usepackage{amsmath,amssymb}
\usepackage{algorithm2e}
\usepackage{enumitem}
\usepackage{wrapfig}
\usepackage{hyperref}
\usepackage{url}
\usepackage{booktabs}
\usepackage{graphicx}
\usepackage[table]{xcolor}
\usepackage{multirow}
\usepackage{wrapfig}
\usepackage{graphicx}
\usepackage{subcaption}
\usepackage{xspace}
\usepackage{comment}
\usepackage{adjustbox}
\usepackage{siunitx}
\usepackage[table]{xcolor}

\definecolor{lightBlue}{HTML}{E5F0FF}

\definecolor{loraOrange}{HTML}{FFF0DF}

\definecolor{fullGreen}{HTML}{E2F3E8}

\usepackage{tikz}
\usepackage{xcolor}
\usepackage{colortbl}
\usepackage{multirow}
\usepackage{booktabs}
\usepackage{adjustbox}

\definecolor{fullFTblue}{HTML}{E5F0FF}
\definecolor{loraOrange}{HTML}{FFF0DF}
\definecolor{lightGreen}{HTML}{E2F3E8}

\definecolor{sqaLoRABg}{HTML}{FFF0DF}
\definecolor{sqaDeltaBg}{HTML}{E5F0FF}
\definecolor{sqaHeaderBg}{HTML}{F2F2F2}
\definecolor{sqaAccent}{HTML}{A3546A}

\newcommand{\method}{\ensuremath{\text{LiT}}\xspace}

\title{\method: Towards Efficient Training for Continual Learning Intelligence}
\logo{\includegraphics[height=12mm]{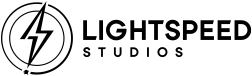}}
\secondarylogos{%
  \includegraphics[height=12mm]{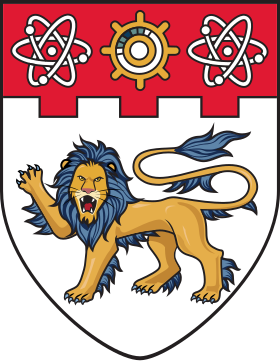}%
}

\definecolor{myblue}{RGB}{224,242,255}
\definecolor{qwenpurple}{HTML}{EEE7FF}  
\definecolor{llavaorange}{HTML}{FFE8CC} 

\author[1]{Jingdi Lei}
\author[2]{Di Zhang}
\author[3]{Zhanqiu Zhang}
\author[4]{Yiwen Guo}
\author[1]{Soujanya Poria}
\affiliation[1]{Nanyang Technological University}
\affiliation[2]{Fudan University}
\affiliation[3]{LIGHTSPEED}
\affiliation[4]{Independent Researcher}

\metadata[{\faGithub\ Github}]{
  \href{https://github.com/declare-lab/LiT}{\textbf{\method}}
}
\metadata[{\faEnvelope\ Correspondence}]{
  \href{jingdi001@e.ntu.edu.sg}{Jingdi Lei}, \href{zqzhang27@gmail.com}{Zhanqiu Zhang}, \href{guoyiwen89@gmail.com}{Yiwen Guo}, \href{soujanya.poria@ntu.edu.sg}{Soujanya Poria}
}

\abstract{
\vspace{0.8em}

As large language models increasingly serve as personal agents, they must adapt to diverse user needs and evolving environments. Continual learning seeks to enable this adaptation by allowing models to progressively acquire new knowledge and capabilities while retaining what they have previously learned. The computational cost of continual adaptation has motivated the widespread adoption of parameter-efficient training methods such as low-rank adaptation~(LoRA). However, sequentially updating a shared LoRA adapter can lead to interference between tasks and catastrophic forgetting, while maintaining task-specific adapters complicates the integration of acquired knowledge into a unified model. This raises a natural question: can parameter-efficient training be achieved by directly updating a small subset of the model's existing weights, without relying on low-rank parameterization? Surprisingly, we find that this is indeed possible. We introduce \textbf{\method}, a simple yet effective training method that directly optimizes a uniformly random subset of model weight coordinates without imposing low-rank constraints. Under matched trainable parameter budgets, \textbf{\method} outperforms LoRA in both continual learning and reinforcement learning settings. On TRACE, \method outperforms LoRA by up to 18.95 points in sequential learning, while reducing performance degradation relative to single-task training by up to 58.2\% when merging independently trained updates. Across both settings, \method also incurs substantially less degradation to general capabilities and safety alignment. In reinforcement learning, \method outperforms both LoRA and full fine-tuning across all six evaluated settings. These results demonstrate that \method provides a more effective alternative to LoRA for the efficient training of continual learning.
}

\date{\today}

\begin{document}
\maketitle

\section{Introduction}
\label{Intro}
Large language models~(LLMs) are extending beyond general purpose assistance to support personalized intelligent systems, requiring them to adapt to the diverse needs, preferences, and evolving environments of individuals~\citep{liu2025personalized}. Such systems must continuously acquire new knowledge and refine their behavior through ongoing interactions, while retaining what they have previously learned~\citep{shi2025continual,mccloskey1989catastrophic}. 

While inference-time approaches such as in-context learning and external memory~\citep{brown2020language,lewis2020rag} can support personalization without modifying weights, they are constrained by context overhead and cannot fundamentally internalize new skills into the model. Continual learning offers a complementary path by progressively incorporating new knowledge and experiences directly into model parameters. For this to be practical, each update must be inexpensive to train while preserving previously acquired knowledge. Low-rank adaptation (LoRA)~\citep{hu2021lora} has become a widely adopted approach to efficient parameter updates. However, sequentially updating a shared LoRA adapter with new tasks can cause interference between successive updates, leading to catastrophic forgetting of previously acquired knowledge~\citep{mccloskey1989catastrophic,kirkpatrick2017overcoming, wang2023olora,liang2024inflora}. An alternative is to maintain a separate adapter for each task, which helps preserve previous adaptations but leaves learned knowledge distributed across multiple adapters. This requires additional mechanisms to select, combine, or consolidate adapters~\citep{pfeiffer2021adapterfusion,huang2023lorahub}, making it difficult to maintain a single model that continuously integrates new experiences.

These limitations motivate a natural question: can parameter-efficient training be achieved by directly optimizing a small subset of the model weights, without relying on low-rank parameterization? Motivated by the idea that sparse subnetworks can retain substantial learning capacity~\citep{frankle2019lottery}, we answer this question by introducing \method, a simple yet effective approach that trains a uniformly random subset of model weight coordinates. Specifically, \method samples a fixed set of coordinates from each projection and learns sparse additive updates at these positions, while keeping the original weights frozen. Unlike LoRA, which represents weight updates through the product of low-rank matrices, \method operates directly in the original weight space without imposing a low-rank constraint or requiring data-dependent parameter selection. This formulation significantly reduces the number of trainable parameters and their associated optimizer states.

We evaluate \method{} across continual learning and reinforcement learning~(RL), comparing it against LoRA under matched trainable parameter budgets. For continual learning, we investigate two settings for integrating knowledge across tasks: sequential learning and independent training followed by merging. On TRACE, after sequentially learning eight tasks, \method{} improved by up to 18.95 points over LoRA while reducing the forgetting of previously learned tasks by up to 91.1\% compared with LoRA. Under independent training and merging, \method{} reduced the performance degradation relative to single-task training by up to 58.2\% compared with LoRA, while effectively preserving capabilities acquired across multiple tasks. Moreover, \method{} better preserved the model's general capabilities and safety-related behaviors than LoRA under both settings, achieving stronger performance on IFEval~\citep{zhou2023ifeval} and HumanEval~\citep{chen2021humaneval}, as well as substantially lower unsafe response rates on AdvBench~\citep{zou2023advbench} and HarmBench~\citep{mazeika2024harmbench}. In reinforcement learning, \method{} achieved higher average accuracy than both LoRA and full fine-tuning in all six evaluated settings across three model scales under GRPO and DAPO. Notably, on Qwen3.5-9B under DAPO, \method{} achieved an average accuracy of 69.33\%, outperforming LoRA and full fine-tuning by 5.35 and 1.47 points, respectively. Furthermore, with FP8 base quantization and W8A16 computation, \method{} enabled reinforcement learning of a 4B-parameter model on a single 24GB RTX 4090D GPU, demonstrating its feasibility on consumer-level hardware.

Our main contributions are summarized as follows:
\begin{itemize}
    \item We introduce \method, a simple yet effective training approach that directly optimizes a uniformly random subset of the model weights. \method substantially reduces the number of trainable parameters and optimization overhead.
    
    \item We show that \method{} mitigates catastrophic forgetting during sequential learning, improves multi-task knowledge integration under model merging, and better preserves general capabilities and safety behaviors compared with LoRA, enabling continuous adaptation within a single unified model.

    \item We demonstrate the effectiveness of \method in RL. Furthermore, \method enables RL training of a 4B model on a single consumer-grade 24GB RTX 4090D GPU.

\end{itemize}

\section{Preliminaries}
\label{sec:preliminary}

\subsection{Low-Rank Adaptation.}
LoRA freezes an original weight matrix $W_0 \in \mathbb{R}^{d_{\mathrm{out}} \times d_{\mathrm{in}}}$ and parameterizes its update via low-rank factorization:
\begin{equation}
    W = W_0 + \Delta W,
    \qquad
    \Delta W = \frac{\gamma}{r} BA,
    \label{eq:lora}
\end{equation}
where $B \in \mathbb{R}^{d_{\mathrm{out}} \times r}$ and $A \in \mathbb{R}^{r \times d_{\mathrm{in}}}$ are trainable matrices with rank $r \ll \min(d_{\mathrm{in}}, d_{\mathrm{out}})$, and $\gamma$ is a scaling factor. This reduces the number of trainable parameters from $d_{\mathrm{out}}d_{\mathrm{in}}$ to $r(d_{\mathrm{in}}+d_{\mathrm{out}})$ by constraining $\Delta W$ to a low-rank subspace, rather than directly optimizing model weights.

\subsection{Continual Learning Setup}

We consider a collection of $T$ tasks $\{\mathcal{T}_1, \mathcal{T}_2, \ldots, \mathcal{T}_T\}$, where each task $\mathcal{T}_t$ is associated with a training dataset $\mathcal{D}_t$. Let $\theta_0$ denote the full set of model parameters. We evaluate two settings for consolidating multi-task knowledge into a single model.

\paragraph{Sequential Learning.}
Tasks are learned sequentially, where each task updates the parameters obtained from the previous stage:
\begin{equation}
    \theta_t = \operatorname{Update}(\theta_{t-1}, \mathcal{D}_t), \quad t = 1, \ldots, T.
\end{equation}
The central challenge is forgetting, where adapting to new tasks degrades performance on earlier ones.

\paragraph{Independent Training and Merging.}
Each task is trained independently from $\theta_0$ to produce a task-specific update $\Delta\theta_t$, and all updates are directly summed into a unified model:
\begin{equation}
    \Delta\theta_t = \operatorname{Update}(\theta_0, \mathcal{D}_t) - \theta_0, \qquad \theta_{\mathrm{merge}} = \theta_0 + \sum_{t=1}^{T}\Delta\theta_t.
\end{equation}
The challenge here is minimizing parameter interference when integrating independently learned updates.

\section{\method}
\label{sec:method}

\paragraph{Sparse Parameter Selection.}

Consider a collection of $M$ weight matrices $\{W_0^{(j)}\}_{j=1}^{M}$, $N_j$ denotes the number of parameters in the $j$-th matrix, and $N=\sum_{j=1}^{M}N_j$ denotes the total number of candidate weights. Given a sampling density $\rho \in (0, 1]$, we randomly select $K = \lfloor \rho N \rfloor$ trainable coordinates across all matrices.

To distribute the coordinate budget proportionally to matrix sizes, we adopt Hamilton's largest-remainder method. For each matrix, we first compute its ideal allocation $a_j=KN_j/N$ and assign $\lfloor a_j\rfloor$ coordinates~($\lfloor \cdot \rfloor$ rounding down to the nearest integer). The remaining $R=K-\sum_{j=1}^{M}\lfloor a_j\rfloor$ coordinates are distributed to the $R$ matrices with the largest fractional remainders. This determines \(K_j\), the number of trainable coordinates allocated to the \(j\)-th matrix, ensuring that \(\sum_{j=1}^{M}K_j=K\). Within each matrix, the allocated coordinates are sampled uniformly without replacement, independently of the training data, gradients, or weight importance, and remain fixed throughout training.

\paragraph{Sparse Weight Update.} For each selected coordinate $(r_k^{(j)}, c_k^{(j)})$, we introduce a trainable scalar $v_k^{(j)}$ that directly updates the corresponding original weight. Specifically, the effective weight matrix is defined as 
\begin{equation}
    W^{(j)}_{r,c} = W^{(j)}_{0, r,c} + \sum_{k=1}^{K_j}v_k^{(j)}\mathbb{I}[(r,c) = (r_k^{(j)}, c_k^{(j)})].
\end{equation}
$\mathbb{I}[\cdot]$ is an indicator function that equals one when the selected coordinate matches $(r,c)$, and zero otherwise. All trainable values are initialized to zero, ensuring that the effective weights are identical to the model weights at initialization. Consequently, only the selected coordinates are modified during training, while all remaining weights stay frozen. 

The resulting weight matrix is used directly in the standard form. Given input activations $X \in \mathbb{R}^{L\times d_{in}}$, the output is computed as 
\begin{equation}
    Y_{\ell,r} = \sum_{c=1}^{d_{in}}X_{\ell,c}W_{r,c},
\end{equation}
where $L$ denotes the number of input tokens. Thus, each selected parameter directly modifies its corresponding entry in the original weight matrix, without introducing additional low-rank factors or a separate low-rank parameterization.

\paragraph{Sparse Gradient Computation.}
Let $\mathcal{L}$ denote the training objective and
$G=\partial\mathcal{L}/\partial Y
\in\mathbb{R}^{L\times d_{\mathrm{out}}}$
denote the upstream gradient. Given the projection $Y=XW^\top$, its differential is
\begin{equation}
    dY=dXW^\top+XdW^\top.
\end{equation}
Applying the chain rule, the differential of the loss can
be written as
\begin{equation}
\begin{aligned}
    d\mathcal{L} = \langle G,dY\rangle_F = \langle GW,dX\rangle_F + \langle G^\top X,dW\rangle_F,
\end{aligned}
\label{eq:delta_gradient_chain}
\end{equation}
where $\langle\cdot,\cdot\rangle_F$ denotes the
Frobenius inner product. This expression describes
the two gradient paths: gradients propagate through
the effective weight matrix to the input activations,
while the weight gradient determines the updates to
the selected coordinates.

Since only the selected coordinates are trainable,
the weight differential satisfies
\begin{equation}
    dW_{r,c}
    =
    \sum_{k=1}^{K}
    \mathbb{I}[(r,c)=(r_k,c_k)]\,dv_k.
\end{equation}
Substituting this expression into
Eq.~\eqref{eq:delta_gradient_chain}, we obtain
\begin{equation}
    d\mathcal{L} = \langle GW,dX\rangle_F+\sum_{k=1}^{K}(G^\top X)_{r_k,c_k}\,dv_k.
\end{equation}
Therefore, the gradient of each trainable parameter is
\begin{equation}
    \frac{\partial\mathcal{L}}{\partial v_k}
    =
    (G^\top X)_{r_k,c_k}
    =
    \sum_{\ell=1}^{L}
    G_{\ell,r_k}X_{\ell,c_k}.
\label{eq:delta_backward}
\end{equation}
This shows that the gradient of each selected weight
is obtained by accumulating the product of its
corresponding input activation and upstream gradient
across all tokens.

Importantly, computing these gradients does not require
constructing the full dense weight-gradient matrix.
We employ sampled dense--dense matrix multiplication
(SDDMM) to evaluate only the $K$ selected entries,
reducing the weight-gradient computation from
$\mathcal{O}(Ld_{\mathrm{out}}d_{\mathrm{in}})$
to $\mathcal{O}(LK)$.

Meanwhile, the gradient propagated to the input
activations is determined by the complete effective
weight matrix:
\begin{equation}
\begin{aligned}
    \frac{\partial\mathcal{L}}{\partial X}=GW=G(W_0+\Delta W).
\end{aligned}
\label{eq:delta_input_gradient}
\end{equation}

The first term propagates gradients through the
original weights, while the second accounts for
the contributions of the selected weight updates.
Thus, gradients are propagated through the entire
network using the effective weights, whereas only
the selected coordinates require weight-gradient
computation and optimizer states.

\section{Experiments}
\label{sec:exp}

\subsection{Continual Learning}
\label{sec:continual_learning}
\paragraph{Models and Baselines.}
We evaluate \method{} on three base models: Qwen3-4B-Base, Qwen3-8B-Base~\citep{qwen3technicalreport}, and Qwen3.5-9B-Base~\citep{qwen35} (referred to as Qwen3-4B, Qwen3-8B, and Qwen3.5-9B hereafter for brevity), across both sequential learning and independent training followed by merging. To evaluate memory retention and continual adaptation under an extreme parameter-efficiency regime, we use LoRA with rank $r=1$~(scaling factor $\gamma=2$) and allocate the exact same number of trainable coordinates to \method{}. We sweep learning rates over $\{3\times10^{-5}, 5\times10^{-5}, 1\times10^{-4}, 2\times10^{-4}, 5\times10^{-4}, 1\times10^{-3}, 2\times10^{-3}\}$ for both methods, each initialized from the same base model with random seed 42. For each model, method, and setting, we select the learning rate that achieves the best performance on the TRACE validation set and report the corresponding results on the test set. More training setting is detailed in Appendix~\ref{app:cl-settings}. During sequential learning, we continuously update the same trainable coordinate set or LoRA adapter across all tasks, resetting optimizer states between tasks. For independent training followed by merging, each task starts from the original base model with an independently sampled coordinate set (for LiT) or adapter initialization (for LoRA), deterministically derived from random seed 42, and the resulting task updates are directly summed into the base weights.

\begin{table*}[htbp]
    \centering
    \caption{
    Sequential learning performance on TRACE after training on all eight tasks across three models. We report task-wise performance after eight tasks training, overall performance~(OP), and backward transfer~(BWT); LR denotes the learning rate.
    }
    \label{tab:sequential_trace}

    \setlength{\tabcolsep}{4pt}
    \renewcommand{\arraystretch}{1.22}

    \resizebox{\textwidth}{!}{%
    \begin{tabular}{
        l l c @{\hspace{8pt}}
        *{8}{c}
        @{\hspace{10pt}} c c
    }
        \toprule

        \textbf{Model}
        & \textbf{Method}
        & \textbf{LR}
        & \textbf{C-STANCE}
        & \textbf{FOMC}
        & \textbf{MeetingBank}
        & \textbf{Py150}
        & \textbf{ScienceQA}
        & \textbf{NumGLUE-cm}
        & \textbf{NumGLUE-ds}
        & \textbf{20Minuten}
        & \textbf{OP $\uparrow$}
        & \textbf{BWT $\uparrow$} \\

        \midrule

        \multirow{2}{*}{\textbf{Qwen3-4B}}
        & \textcolor{black!75}{LoRA}
        & \textcolor{black!75}{$5\!\times\!10^{-5}$}
        & \textbf{52.15} & 54.84 & 29.69 & 63.02
        & 84.95 & \textbf{70.37} & \textbf{67.08} & 39.98
        & 57.76 & $-5.82$ \\

        & \cellcolor{sqaDeltaBg}\textbf{\method{}}
        & \cellcolor{sqaDeltaBg}$5\!\times\!10^{-4}$
        & \cellcolor{sqaDeltaBg}51.95 & \cellcolor{sqaDeltaBg}\textbf{66.94} & \cellcolor{sqaDeltaBg}\textbf{53.62} & \cellcolor{sqaDeltaBg}\textbf{67.00}
        & \cellcolor{sqaDeltaBg}\textbf{87.40} & \cellcolor{sqaDeltaBg}66.67 & \cellcolor{sqaDeltaBg}63.38 & \cellcolor{sqaDeltaBg}\textbf{40.25}
        & \cellcolor{sqaDeltaBg}\textbf{62.15} & \cellcolor{sqaDeltaBg}$\mathbf{-1.87}$ \\

        \cmidrule(lr){1-13}

        \multirow{2}{*}{\textbf{Qwen3-8B}}
        & \textcolor{black!75}{LoRA}
        & \textcolor{black!75}{$5\!\times\!10^{-4}$}
        & \textbf{55.65} & 63.10 & 15.02 & 59.34
        & 85.30 & 67.90 & \textbf{72.00} & \textbf{41.61}
        & 57.49 & $-10.83$ \\

        & \cellcolor{sqaDeltaBg}\textbf{\method{}}
        & \cellcolor{sqaDeltaBg}$1\!\times\!10^{-3}$
        & \cellcolor{sqaDeltaBg}55.45 & \cellcolor{sqaDeltaBg}\textbf{68.55} & \cellcolor{sqaDeltaBg}\textbf{56.55} & \cellcolor{sqaDeltaBg}\textbf{67.06}
        & \cellcolor{sqaDeltaBg}\textbf{91.00} & \cellcolor{sqaDeltaBg}\textbf{77.78} & \cellcolor{sqaDeltaBg}70.46 & \cellcolor{sqaDeltaBg}40.90
        & \cellcolor{sqaDeltaBg}\textbf{65.97} & \cellcolor{sqaDeltaBg}$\mathbf{-1.25}$ \\

        \cmidrule(lr){1-13}

        \multirow{2}{*}{\textbf{Qwen3.5-9B}}
        & \textcolor{black!75}{LoRA}
        & \textcolor{black!75}{$2\!\times\!10^{-4}$}
        & 23.20 & 0.81 & 16.34 & 57.82
        & 94.45 & \textbf{77.78} & \textbf{74.77} & \textbf{42.07}
        & 48.40 & $-20.32$ \\

        & \cellcolor{sqaDeltaBg}\textbf{\method{}}
        & \cellcolor{sqaDeltaBg}$1\!\times\!10^{-3}$
        & \cellcolor{sqaDeltaBg}\textbf{59.30} & \cellcolor{sqaDeltaBg}\textbf{66.94} & \cellcolor{sqaDeltaBg}\textbf{57.90} & \cellcolor{sqaDeltaBg}\textbf{66.50}
        & \cellcolor{sqaDeltaBg}\textbf{94.85} & \cellcolor{sqaDeltaBg}76.54 & \cellcolor{sqaDeltaBg}\textbf{74.77} & \cellcolor{sqaDeltaBg}41.97
        & \cellcolor{sqaDeltaBg}\textbf{67.35} & \cellcolor{sqaDeltaBg}$\mathbf{-1.80}$ \\

        \bottomrule
    \end{tabular}%
    }
\end{table*}

\paragraph{Benchmarks and Metrics.}
We evaluate continual learning on TRACE~\citep{wang2023trace} across its eight tasks using their official evaluation metrics. For sequential learning, we report overall performance~(OP) and backward transfer~(BWT) to measure the final multi-task performance and the extent of forgetting across previously learned tasks. For independent training followed by merging, we report average task performance and negative log-likelihood~(NLL) before and after merging. To directly probe factual knowledge retention, we additionally conduct a retention test on SimpleQA~\citep{wei2024simpleqa}: we first train each model on 100 SimpleQA examples for 100 epochs until full memorization, and then continue training on the shuffled TRACE training set while tracking accuracy on the memorized SimpleQA questions. We further evaluate the preservation of general capabilities after both sequential learning and merging on TRACE using IFEval~\citep{zhou2023ifeval} for instruction following and HumanEval~\citep{chen2021humaneval} for code generation. To assess safety preservation, we use Qwen3Guard-Gen-8B~\citep{zhao2025qwen3guard} to evaluate unsafe response rates and refusal rates on AdvBench~\citep{zou2023advbench} and HarmBench~\citep{mazeika2024harmbench}. 

\vspace{-4mm}

\paragraph{Sequential Learning.}
\begin{table*}[htbp]
    \centering
    \caption{
    Task-wise performance comparison between LoRA and \method{} on TRACE under single-task and merging settings across three models. Higher task scores and lower NLL indicate better performance.
    }
    \label{tab:trace_combine_main}

    \setlength{\tabcolsep}{2.5pt}
    \renewcommand{\arraystretch}{1.}

    \resizebox{\textwidth}{!}{%
    \begin{tabular}{
        l l @{\hspace{8pt}}
        *{8}{c}
        @{\hspace{10pt}} c c
    }
        \toprule

        \textbf{Method}
        & \textbf{Setting}
        & \textbf{C-STANCE}
        & \textbf{FOMC}
        & \textbf{MeetingBank}
        & \textbf{Py150}
        & \textbf{ScienceQA}
        & \textbf{NumGLUE-cm}
        & \textbf{NumGLUE-ds}
        & \textbf{20Minuten}
        & \textbf{Avg. $\uparrow$}
        & \textbf{NLL $\downarrow$} \\

        \midrule
        \addlinespace[2pt]

        \rowcolor{gray!10}
        \multicolumn{12}{l}{
            \textbf{\textit{Qwen3-4B}}
            \quad\footnotesize\textcolor{gray!80!black}{(LR: LoRA $5\!\times\!10^{-5}$, \method{} $5\!\times\!10^{-4}$)}
        } \\
        \addlinespace[2pt]

        & \textcolor{black!60}{Single-task}
        & \textcolor{black!60}{55.60} & \textcolor{black!60}{64.52} & \textcolor{black!60}{58.50} & \textcolor{black!60}{66.55}
        & \textcolor{black!60}{86.50} & \textcolor{black!60}{71.60} & \textcolor{black!60}{66.77} & \textcolor{black!60}{39.73}
        & \textcolor{black!60}{63.72}
        & \textcolor{black!60}{0.993} \\

        \multirow{-2}{*}{\textbf{LoRA}}
        & \textbf{Merging}
        & 51.50 & 58.06 & 52.88 & 64.46
        & 86.80 & 66.67 & 59.38 & 40.57
        & 60.04\,{\scriptsize\textcolor{gray!80!black}{($\downarrow$3.68)}}
        & 1.270\,{\scriptsize\textcolor{gray!80!black}{($\uparrow$0.276)}} \\

        \addlinespace[2pt]
        \cmidrule(lr){2-12}
        \addlinespace[2pt]

        & \textcolor{black!60}{Single-task}
        & \textcolor{black!60}{53.60} & \textcolor{black!60}{68.95} & \textcolor{black!60}{57.84} & \textcolor{black!60}{67.54}
        & \textcolor{black!60}{87.85} & \textcolor{black!60}{72.84} & \textcolor{black!60}{66.46} & \textcolor{black!60}{39.73}
        & \textcolor{black!60}{64.35}
        & \textcolor{black!60}{0.972} \\

        \rowcolor{sqaDeltaBg}
        \cellcolor{white}\multirow{-2}{*}{\textbf{\method{}}}
        & \textbf{Merging}
        & 53.30 & 64.72 & 52.58 & 66.53
        & 86.25 & 62.96 & 61.23 & 40.78
        & \textbf{61.04}\,{\scriptsize\textcolor{gray!80!black}{($\downarrow$3.31)}}
        & \textbf{1.211}\,{\scriptsize\textcolor{gray!80!black}{($\uparrow$0.240)}} \\

        \addlinespace[4pt]
        \midrule
        \addlinespace[4pt]

        \rowcolor{gray!10}
        \multicolumn{12}{l}{
            \textbf{\textit{Qwen3-8B}}
            \quad\footnotesize\textcolor{gray!80!black}{(LR: LoRA $5\!\times\!10^{-4}$, \method{} $2\!\times\!10^{-3}$)}
        } \\
        \addlinespace[2pt]

        & \textcolor{black!60}{Single-task}
        & \textcolor{black!60}{57.25} & \textcolor{black!60}{72.78} & \textcolor{black!60}{63.93} & \textcolor{black!60}{66.29}
        & \textcolor{black!60}{93.85} & \textcolor{black!60}{75.31} & \textcolor{black!60}{78.77} & \textcolor{black!60}{41.33}
        & \textcolor{black!60}{68.69}
        & \textcolor{black!60}{1.058} \\

        \multirow{-2}{*}{\textbf{LoRA}}
        & \textbf{Merging}
        & 51.75 & 62.30 & 33.26 & 61.80
        & 82.85 & 69.14 & 63.08 & 40.60
        & 58.10\,{\scriptsize\textcolor{gray!80!black}{($\downarrow$10.59)}}
        & 3.034\,{\scriptsize\textcolor{gray!80!black}{($\uparrow$1.976)}} \\

        \addlinespace[2pt]
        \cmidrule(lr){2-12}
        \addlinespace[2pt]

        & \textcolor{black!60}{Single-task}
        & \textcolor{black!60}{58.55} & \textcolor{black!60}{68.95} & \textcolor{black!60}{61.77} & \textcolor{black!60}{68.16}
        & \textcolor{black!60}{91.50} & \textcolor{black!60}{76.54} & \textcolor{black!60}{73.23} & \textcolor{black!60}{41.59}
        & \textcolor{black!60}{67.54}
        & \textcolor{black!60}{0.904} \\

        \rowcolor{sqaDeltaBg}
        \cellcolor{white}\multirow{-2}{*}{\textbf{\method{}}}
        & \textbf{Merging}
        & 58.05 & 64.52 & 47.47 & 64.14
        & 89.75 & 71.60 & 68.31 & 41.08
        & \textbf{63.11}\,{\scriptsize\textcolor{gray!80!black}{($\downarrow$4.43)}}
        & \textbf{1.307}\,{\scriptsize\textcolor{gray!80!black}{($\uparrow$0.403)}} \\

        \addlinespace[4pt]
        \midrule
        \addlinespace[4pt]

        \rowcolor{gray!10}
        \multicolumn{12}{l}{
            \textbf{\textit{Qwen3.5-9B}}
            \quad\footnotesize\textcolor{gray!80!black}{(LR: LoRA $5\!\times\!10^{-5}$, \method{} $5\!\times\!10^{-4}$)}
        } \\
        \addlinespace[2pt]

        & \textcolor{black!60}{Single-task}
        & \textcolor{black!60}{60.35} & \textcolor{black!60}{69.76} & \textcolor{black!60}{62.25} & \textcolor{black!60}{67.23}
        & \textcolor{black!60}{93.65} & \textcolor{black!60}{79.01} & \textcolor{black!60}{75.69} & \textcolor{black!60}{41.64}
        & \textcolor{black!60}{68.70}
        & \textcolor{black!60}{0.821} \\

        \multirow{-2}{*}{\textbf{LoRA}}
        & \textbf{Merging}
        & 56.35 & 66.33 & 58.75 & 66.59
        & 93.35 & 70.37 & 67.69 & 41.33
        & 65.10\,{\scriptsize\textcolor{gray!80!black}{($\downarrow$3.60)}}
        & 1.078\,{\scriptsize\textcolor{gray!80!black}{($\uparrow$0.257)}} \\

        \addlinespace[2pt]
        \cmidrule(lr){2-12}
        \addlinespace[2pt]

        & \textcolor{black!60}{Single-task}
        & \textcolor{black!60}{60.35} & \textcolor{black!60}{68.75} & \textcolor{black!60}{60.82} & \textcolor{black!60}{67.17}
        & \textcolor{black!60}{93.75} & \textcolor{black!60}{80.25} & \textcolor{black!60}{75.38} & \textcolor{black!60}{41.33}
        & \textcolor{black!60}{68.47}
        & \textcolor{black!60}{0.805} \\

        \rowcolor{sqaDeltaBg}
        \cellcolor{white}\multirow{-2}{*}{\textbf{\method{}}}
        & \textbf{Merging}
        & 58.10 & 70.16 & 58.43 & 67.20
        & 93.35 & 74.07 & 69.54 & 40.77
        & \textbf{66.45}\,{\scriptsize\textcolor{gray!80!black}{($\downarrow$2.02)}}
        & \textbf{0.975}\,{\scriptsize\textcolor{gray!80!black}{($\uparrow$0.171)}} \\

        \addlinespace[2pt]
        \bottomrule
    \end{tabular}%
    }
\end{table*}
Table~\ref{tab:sequential_trace} presents the results of sequentially learning eight tasks on TRACE across all three models. Even when continuously updating the same fixed set of sparse coordinates across all tasks, \method{} consistently outperforms LoRA in both overall performance~(OP) and backward transfer~(BWT) at every model scale. Specifically, \method{} achieves an OP of 62.15 on Qwen3-4B, 65.97 on Qwen3-8B, and 67.35 on Qwen3.5-9B, compared with 57.76, 57.49, and 48.40 for LoRA. Across the same three models, \method{} keeps BWT tightly bounded between $-1.25$ and $-1.87$, whereas LoRA degrades to $-5.82$, $-10.83$, and $-20.32$, respectively. Notably, while increasingly severe forgetting causes LoRA's OP to drop on larger models, \method{} scales monotonically with model size. Task-wise, both methods attain comparable accuracy on the most recently learned tasks such as NumGLUE-ds and 20Minuten, but LoRA suffers severe collapse on earlier tasks---most strikingly on MeetingBank across all three models, as well as FOMC and C-STANCE on Qwen3.5-9B. These results demonstrate that \method{} acquires new task knowledge with substantially less overwriting of previously learned capabilities than LoRA.

\paragraph{Independent Training and Merging.}
Table~\ref{tab:trace_combine_main} evaluates direct merging of independently trained task updates on TRACE. Across all three model scales, \method{} achieves higher merged average accuracy and lower post-merge NLL than LoRA, while incurring smaller drops relative to single-task training. Specifically, \method{} outperforms LoRA in merged accuracy by 1.00 point on Qwen3-4B, 5.01 points on Qwen3-8B, and 1.35 points on Qwen3.5-9B, reaching 66.45 at the 9B scale. The contrast is starkest on Qwen3-8B and Qwen3.5-9B: although LoRA attains slightly higher single-task averages, direct merging causes much sharper losses in both accuracy and likelihood. On Qwen3-8B, merging degrades LoRA by 10.59 points and drives its NLL from 1.058 to 3.034, whereas \method{} drops by only 4.43 points, or 58.2\% less degradation, with NLL rising modestly from 0.904 to 1.307. These results confirm that strong single-task adaptation does not automatically translate into effective multi-task integration, whereas random sparse coordinate updates exhibit substantially less cross-task interference when combined.

\paragraph{General Capability and Safety Preservation.}
Table~\ref{tab:general_preservation} evaluates the preservation of general capabilities and safety alignment after training on TRACE. Across both sequential learning and merging, \method{} outperforms LoRA on every general capability metric and stays much closer to the vanilla baseline. The contrast is sharpest under merging on Qwen3-8B: starting from vanilla scores of 64.63 on HumanEval and 43.81 on IFEval, LoRA collapses to 7.32 and 12.38, whereas \method{} retains 59.15 and 32.16. On Qwen3-4B and Qwen3.5-9B, \method{} likewise remains close to the base models across both settings, reaching 72.27 on Qwen3.5-9B IFEval after sequential learning, compared with 73.01 for vanilla and 49.72 for LoRA. For safety preservation, \method{} surpasses LoRA in 23 of 24 comparisons across AdvBench and HarmBench. Under sequential learning, LoRA induces severe safety erosion on Qwen3-4B and Qwen3-8B, pushing AdvBench unsafe rates from 18.27\% and 8.65\% up to 67.31\% and 51.73\% while dropping refusal rates to 44.62\% and 57.50\%. In contrast, \method{} keeps unsafe rates near the vanilla baselines at 19.04\% and 11.15\%, with refusal rates of 82.69\% and 91.73\%, and exhibits a similar margin on HarmBench. These results show that \method{} preserves general capabilities and safety alignment far closer to the base models than LoRA during continual adaptation.
\begin{table*}[t]
    \centering
    \caption{
    Preservation of general capabilities and safety-related behaviors after integrating eight tasks on TRACE across three models under independent training followed by merging and sequential learning (Van. denotes Vanilla, and LR denotes the selected learning rate). Higher IFEval, HumanEval, and refusal rates ($\uparrow$) and lower unsafe response rates ($\downarrow$) indicate better performance. Unsafe and refusal rates are measured independently.
    }
    \label{tab:general_preservation}

    \setlength{\tabcolsep}{4pt}
    \renewcommand{\arraystretch}{1.18}

    \resizebox{\textwidth}{!}{%
    \begin{tabular}{
        l @{\hspace{10pt}}
        c @{\hspace{8pt}} c >{\columncolor{sqaDeltaBg}}c @{\hspace{8pt}} c >{\columncolor{sqaDeltaBg}}c
        @{\hspace{14pt}}
        c @{\hspace{8pt}} c >{\columncolor{sqaDeltaBg}}c @{\hspace{8pt}} c >{\columncolor{sqaDeltaBg}}c
        @{\hspace{14pt}}
        c @{\hspace{8pt}} c >{\columncolor{sqaDeltaBg}}c @{\hspace{8pt}} c >{\columncolor{sqaDeltaBg}}c
    }
        \toprule

        \multirow{3}{*}{\textbf{Metric}}
        & \multicolumn{5}{c}{\textbf{Qwen3-4B}}
        & \multicolumn{5}{c}{\textbf{Qwen3-8B}}
        & \multicolumn{5}{c}{\textbf{Qwen3.5-9B}} \\

        \cmidrule(lr){2-6}
        \cmidrule(lr){7-11}
        \cmidrule(lr){12-16}

        & \multirow{2}{*}{Van.}
        & \multicolumn{2}{c}{\textbf{Merging}}
        & \multicolumn{2}{c}{\textbf{Sequential}}
        & \multirow{2}{*}{Van.}
        & \multicolumn{2}{c}{\textbf{Merging}}
        & \multicolumn{2}{c}{\textbf{Sequential}}
        & \multirow{2}{*}{Van.}
        & \multicolumn{2}{c}{\textbf{Merging}}
        & \multicolumn{2}{c}{\textbf{Sequential}} \\

        \cmidrule(lr){3-4} \cmidrule(lr){5-6}
        \cmidrule(lr){8-9} \cmidrule(lr){10-11}
        \cmidrule(lr){13-14} \cmidrule(lr){15-16}

        & & LoRA & \textbf{\method{}}
        & LoRA & \textbf{\method{}}
        & & LoRA & \textbf{\method{}}
        & LoRA & \textbf{\method{}}
        & & LoRA & \textbf{\method{}}
        & LoRA & \textbf{\method{}} \\

        \midrule

        \rowcolor{gray!10}
        \multicolumn{16}{l}{\textbf{\textit{General Capability Preservation}}} \\

        \hspace{0.6em}IFEval $\uparrow$
        & 38.45 & 21.44 & \textbf{36.97}
        & 23.48 & \textbf{36.04}
        & 43.81 & 12.38 & \textbf{32.16}
        & 24.95 & \textbf{38.08}
        & 73.01 & 61.00 & \textbf{73.20}
        & 49.72 & \textbf{72.27} \\

        \hspace{0.6em}HumanEval $\uparrow$
        & 56.71 & 51.83 & \textbf{59.76}
        & 57.32 & \textbf{60.37}
        & 64.63 & 7.32 & \textbf{59.15}
        & 55.49 & \textbf{60.37}
        & 58.54 & 47.56 & \textbf{56.71}
        & 43.90 & \textbf{55.49} \\

        \midrule

        \rowcolor{gray!10}
        \multicolumn{16}{l}{\textbf{\textit{Safety Preservation}}} \\

        \hspace{0.6em}Adv. Unsafe $\downarrow$
        & 18.27 & 54.42 & \textbf{17.69}
        & 67.31 & \textbf{19.04}
        & 8.65 & 69.62 & \textbf{13.85}
        & 51.73 & \textbf{11.15}
        & 1.54 & \textbf{0.96} & 1.15
        & 8.46 & \textbf{1.73} \\

        \hspace{0.6em}Adv. Refusal $\uparrow$
        & 85.38 & 64.04 & \textbf{85.00}
        & 44.62 & \textbf{82.69}
        & 93.08 & 68.65 & \textbf{88.08}
        & 57.50 & \textbf{91.73}
        & 97.88 & 98.27 & \textbf{98.46}
        & 91.35 & \textbf{97.69} \\

        \hspace{0.6em}Harm. Unsafe $\downarrow$
        & 52.81 & 61.88 & \textbf{53.44}
        & 75.94 & \textbf{51.56}
        & 37.81 & 64.06 & \textbf{41.25}
        & 58.44 & \textbf{35.94}
        & 37.50 & 36.25 & \textbf{35.00}
        & 51.56 & \textbf{39.38} \\

        \hspace{0.6em}Harm. Refusal $\uparrow$
        & 53.75 & 47.50 & \textbf{51.88}
        & 32.81 & \textbf{52.81}
        & 55.31 & 49.06 & \textbf{58.13}
        & 47.50 & \textbf{61.88}
        & 57.19 & 59.06 & \textbf{60.31}
        & 40.62 & \textbf{54.06} \\

        \midrule

        \rowcolor{white}
        \footnotesize\textcolor{gray!80!black}{\textit{Selected LR}}
        & \footnotesize\textcolor{gray!80!black}{--}
        & \footnotesize\textcolor{gray!80!black}{$5\!\times\!10^{-5}$} & \footnotesize\textcolor{gray!80!black}{$5\!\times\!10^{-4}$}
        & \footnotesize\textcolor{gray!80!black}{$5\!\times\!10^{-5}$} & \footnotesize\textcolor{gray!80!black}{$5\!\times\!10^{-4}$}
        & \footnotesize\textcolor{gray!80!black}{--}
        & \footnotesize\textcolor{gray!80!black}{$5\!\times\!10^{-4}$} & \footnotesize\textcolor{gray!80!black}{$2\!\times\!10^{-3}$}
        & \footnotesize\textcolor{gray!80!black}{$5\!\times\!10^{-4}$} & \footnotesize\textcolor{gray!80!black}{$1\!\times\!10^{-3}$}
        & \footnotesize\textcolor{gray!80!black}{--}
        & \footnotesize\textcolor{gray!80!black}{$5\!\times\!10^{-5}$} & \footnotesize\textcolor{gray!80!black}{$5\!\times\!10^{-4}$}
        & \footnotesize\textcolor{gray!80!black}{$2\!\times\!10^{-4}$} & \footnotesize\textcolor{gray!80!black}{$1\!\times\!10^{-3}$} \\

        \bottomrule
    \end{tabular}%
    }
\end{table*}

\begin{figure*}[htbp]
    \centering
    \includegraphics[
        width=0.8\textwidth
    ]{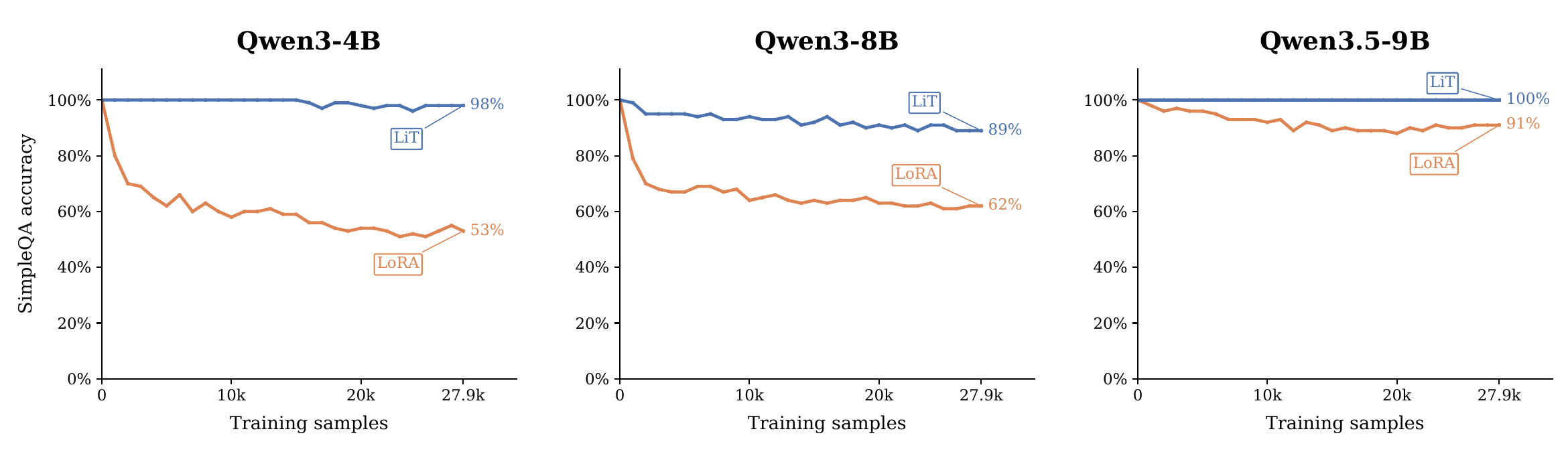}
    \caption{
    Knowledge retention during sequential learning on SimpleQA
    across three models. The horizontal axis represents the number of TRACE training samples, and the vertical axis shows the accuracy on previously learned SimpleQA questions.
    }
    \label{fig:simpleqa_retention}
\end{figure*}

\paragraph{Factual Knowledge Retention.}
Figure~\ref{fig:simpleqa_retention} probes factual forgetting by tracking accuracy on 100 memorized SimpleQA questions during subsequent training on TRACE dataset, reusing the same fixed coordinate set for \method{} and the same adapter for LoRA across both stages. LoRA exhibits rapid factual degradation once subsequent training begins, with accuracy dropping to 53\% on Qwen3-4B, 62\% on Qwen3-8B, and 91\% on Qwen3.5-9B by the end of training. In contrast, \method{} retains 98\%, 89\%, and 100\% accuracy, respectively. This gap confirms that even when updating the exact same coordinates across stages, \method retains previously memorized facts with substantially less interference than LoRA under sustained training.

\paragraph{Effect of Trainable Parameter Budget.}
We sweep the trainable parameter ratio $p$ of \method{} from $10^{-7}$ to $0.5$ on sequential TRACE using Qwen3-8B, scaling the learning rate with budget as detailed in Appendix~\ref{app:lr_scaling}. As shown in Figure~\ref{fig:trace_parameter_budget}, both OP and BWT rise rapidly from the extreme-sparsity regime, and plateau across a broad intermediate window. Notably, LoRA hits its structural minimum at rank $r=1$, which can only compress trainable parameters down to $p = 3.33\times 10^{-4}$. Without a discrete rank floor, \method{} surpasses full fine-tuning across $p \in [3.33\times 10^{-4}, 10^{-1}]$ and remains competitive down to $p = 10^{-4}$, compressing trainable parameters to as little as $0.01\%$---over $3\times$ smaller than rank-1 LoRA and roughly $10{,}000\times$ smaller than full fine-tuning.
\begin{figure*}[htbp]
    \centering
    \includegraphics[width=0.7\textwidth]{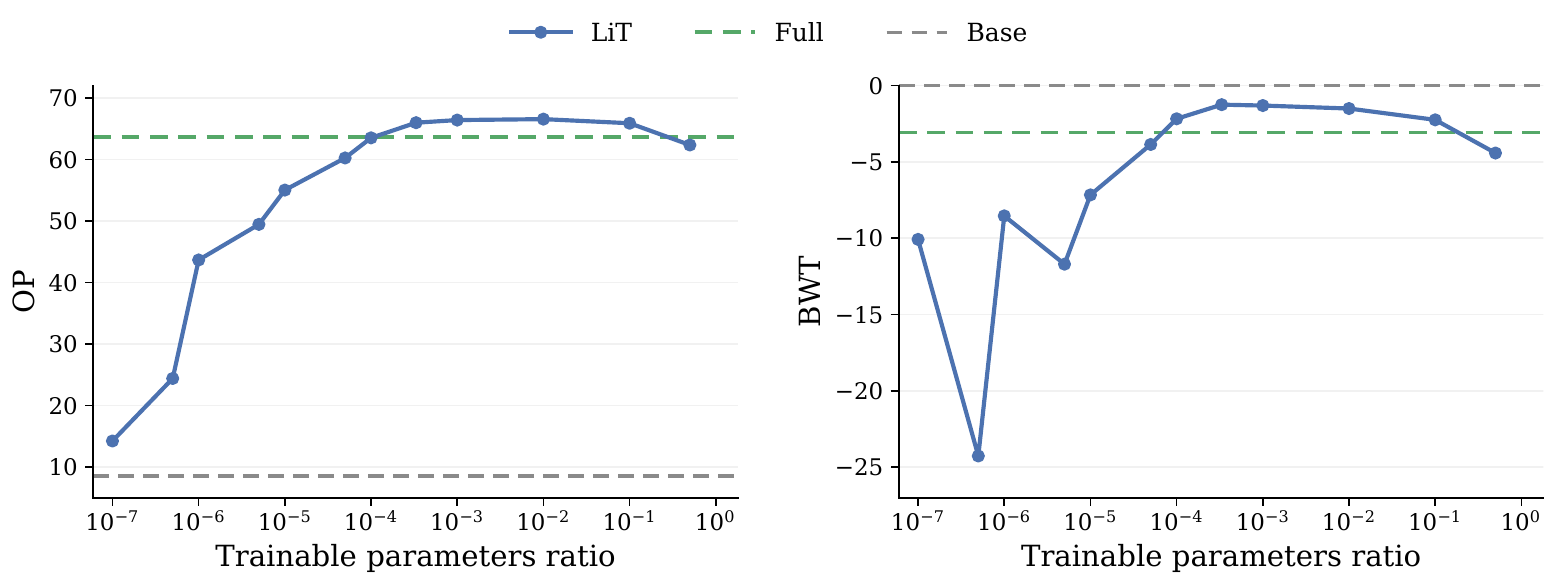}
    \caption{
    Effect of trainable parameter ratio $p$ on sequential learning on TRACE (Qwen3-8B), reporting final OP (left) and BWT (right).
    }
    \label{fig:trace_parameter_budget}
\end{figure*}

\subsection{Reinforcement Learning}
\label{sec:reinforcement_learning}
\paragraph{Models and Baselines.}
We evaluate \method{} on Qwen3-4B~\citep{qwen3technicalreport}, Qwen3-8B~\citep{qwen3technicalreport}, and Qwen3.5-9B~\citep{qwen35}, comparing it against full fine-tuning and LoRA under both GRPO~\citep{shao2024deepseekmath} and DAPO~\citep{yu2025dapo}. For parameter-matched comparisons, we use LoRA with rank $r=16$ and scaling factor $\gamma=32$, and adjust the number of trainable coordinates in \method{} to match the corresponding LoRA parameter budget. In both methods, trainable parameters are restricted to attention and MLP projections.

\paragraph{Training Setup and Evaluation.}
Following~\citet{liu2026unpacking}, we set the learning rate to $2\times10^{-6}$ for full fine-tuning, and $2\times10^{-4}$ for LoRA and \method{}. We evaluate on six mathematical reasoning benchmarks: MATH500~\citep{hendrycks2021math,lightman2023verify}, AIME 2025~\citep{maa2025amc}, AMC23~\citep{maa2025amc}, Minerva~\citep{lewkowycz2022minerva}, OlympiadBench~\citep{he2024olympiadbench}, and College Math~\citep{tang2024mathscale}. We provided more details about training setup and evaluation in Appendix~\ref{app:rl-settings}.

\paragraph{Overall RL Performance.}
\begin{table*}[t]
    \centering
    \caption{
    Reinforcement learning performance across different model scales. We report results on six mathematical reasoning benchmarks and average performance~(Avg.), Van. denotes Vanilla.
    }
    \label{tab:main_rl_performance}

    \setlength{\tabcolsep}{4.5pt}
    \renewcommand{\arraystretch}{1.4}

    \resizebox{\textwidth}{!}{%
    \begin{tabular}{
        l @{\hspace{12pt}}
        c @{\hspace{10pt}} c c >{\columncolor{lightBlue}}c @{\hspace{10pt}} c c >{\columncolor{lightBlue}}c
        @{\hspace{16pt}}
        c @{\hspace{10pt}} c c >{\columncolor{lightBlue}}c @{\hspace{10pt}} c c >{\columncolor{lightBlue}}c
        @{\hspace{16pt}}
        c @{\hspace{10pt}} c c >{\columncolor{lightBlue}}c @{\hspace{10pt}} c c >{\columncolor{lightBlue}}c
    }
        \toprule
        \multirow{3}{*}{\textbf{Benchmark}}
        & \multicolumn{7}{c}{\textbf{Qwen3-4B}}
        & \multicolumn{7}{c}{\textbf{Qwen3-8B}}
        & \multicolumn{7}{c}{\textbf{Qwen3.5-9B}} \\
        \cmidrule(lr){2-8} \cmidrule(lr){9-15} \cmidrule(lr){16-22}

        & \multirow{2}{*}{Van.}
        & \multicolumn{3}{c}{\textbf{GRPO}}
        & \multicolumn{3}{c}{\textbf{DAPO}}
        & \multirow{2}{*}{Van.}
        & \multicolumn{3}{c}{\textbf{GRPO}}
        & \multicolumn{3}{c}{\textbf{DAPO}}
        & \multirow{2}{*}{Van.}
        & \multicolumn{3}{c}{\textbf{GRPO}}
        & \multicolumn{3}{c}{\textbf{DAPO}} \\
        \cmidrule(lr){3-5} \cmidrule(lr){6-8}
        \cmidrule(lr){10-12} \cmidrule(lr){13-15}
        \cmidrule(lr){17-19} \cmidrule(lr){20-22}

        & & Full & LoRA & \textbf{\method} & Full & LoRA & \textbf{\method}
        & & Full & LoRA & \textbf{\method} & Full & LoRA & \textbf{\method}
        & & Full & LoRA & \textbf{\method} & Full & LoRA & \textbf{\method} \\
        \midrule

        \textbf{MATH500}
        & 48.80 & 79.80 & 76.20 & \textbf{83.00} & \textbf{85.20} & 80.20 & 85.00
        & 62.20 & 82.00 & 82.60 & \textbf{84.00} & \textbf{86.40} & 84.60 & 86.00
        & 90.80 & 94.60 & 94.20 & \textbf{94.80} & 96.00 & 94.00 & \textbf{96.80} \\

        \textbf{AIME25}
        & 3.33 & \textbf{18.96} & 18.12 & \textbf{18.96} & 18.75 & 13.96 & \textbf{18.96}
        & 8.33 & 18.12 & 18.12 & \textbf{19.58} & 19.38 & 12.50 & \textbf{22.50}
        & 49.17 & 56.25 & 53.75 & \textbf{60.42} & 60.83 & 50.00 & \textbf{66.25} \\

        \textbf{AMC23}
        & 33.28 & 55.94 & 54.53 & \textbf{59.22} & 62.97 & 60.78 & \textbf{70.94}
        & 42.34 & 63.59 & 63.12 & \textbf{64.06} & 67.97 & 59.84 & \textbf{72.34}
        & 90.16 & 91.56 & 92.03 & \textbf{92.81} & 94.69 & 92.03 & \textbf{97.03} \\

        \textbf{Minerva}
        & 20.59 & 28.68 & 20.59 & \textbf{30.88} & \textbf{33.82} & 25.37 & 33.46
        & 19.49 & 32.35 & 30.51 & \textbf{35.29} & 34.19 & 33.09 & \textbf{34.93}
        & 33.82 & 36.76 & \textbf{40.44} & 38.24 & \textbf{40.44} & 40.07 & 40.07 \\

        \textbf{Olympiad}
        & 26.52 & 44.15 & 38.37 & \textbf{44.44} & \textbf{48.59} & 42.96 & 47.26
        & 28.59 & \textbf{49.19} & 45.93 & 46.96 & 50.07 & 44.89 & \textbf{50.37}
        & 58.81 & 61.04 & \textbf{62.96} & 61.48 & 68.30 & 63.85 & \textbf{69.04} \\

        \textbf{College}
        & 27.50 & 39.53 & 35.27 & \textbf{40.45} & 41.45 & \textbf{42.02} & 41.59
        & 32.33 & \textbf{41.31} & 40.06 & 40.67 & 41.55 & \textbf{43.29} & 41.20
        & 43.01 & 45.53 & \textbf{45.56} & 44.61 & \textbf{46.88} & 43.93 & 46.77 \\

        \midrule
        \rowcolor{gray!10}
        \textbf{Avg.}
        & 26.67 & 44.51 & 40.51 & \cellcolor{lightBlue}\textbf{46.16} & 48.46 & 44.21 & \cellcolor{lightBlue}\textbf{49.53}
        & 32.21 & 47.76 & 46.73 & \cellcolor{lightBlue}\textbf{48.43} & 49.93 & 46.37 & \cellcolor{lightBlue}\textbf{51.22}
        & 60.96 & 64.29 & 64.82 & \cellcolor{lightBlue}\textbf{65.39} & 67.86 & 63.98 & \cellcolor{lightBlue}\textbf{69.33} \\

        \bottomrule
    \end{tabular}%
    }
\end{table*}
As shown in Table~\ref{tab:main_rl_performance}, \method achieves the highest average accuracy across all six evaluated settings spanning three model scales and two RL algorithms. Under GRPO, \method consistently outperforms both full fine-tuning and LoRA, reaching 46.16 on Qwen3-4B compared with 44.51 for full fine-tuning and 40.51 for LoRA, 48.43 on Qwen3-8B compared with 47.76 and 46.73, and 65.39 on Qwen3.5-9B compared with 64.29 and 64.82. The same trend holds under DAPO, where \method attains 49.53 on Qwen3-4B against 48.46 for full fine-tuning and 44.21 for LoRA, 51.22 on Qwen3-8B against 49.93 and 46.37, and 69.33 on Qwen3.5-9B against 67.86 and 63.98. Together, these results demonstrate that directly optimizing a small subset of weight coordinates consistently surpasses both LoRA and full fine-tuning in RL without relying on low-rank parameterization.

\paragraph{Memory and Compute Efficiency.}
\begin{figure*}[htbp]
    \centering
    \includegraphics[width=0.8\textwidth]{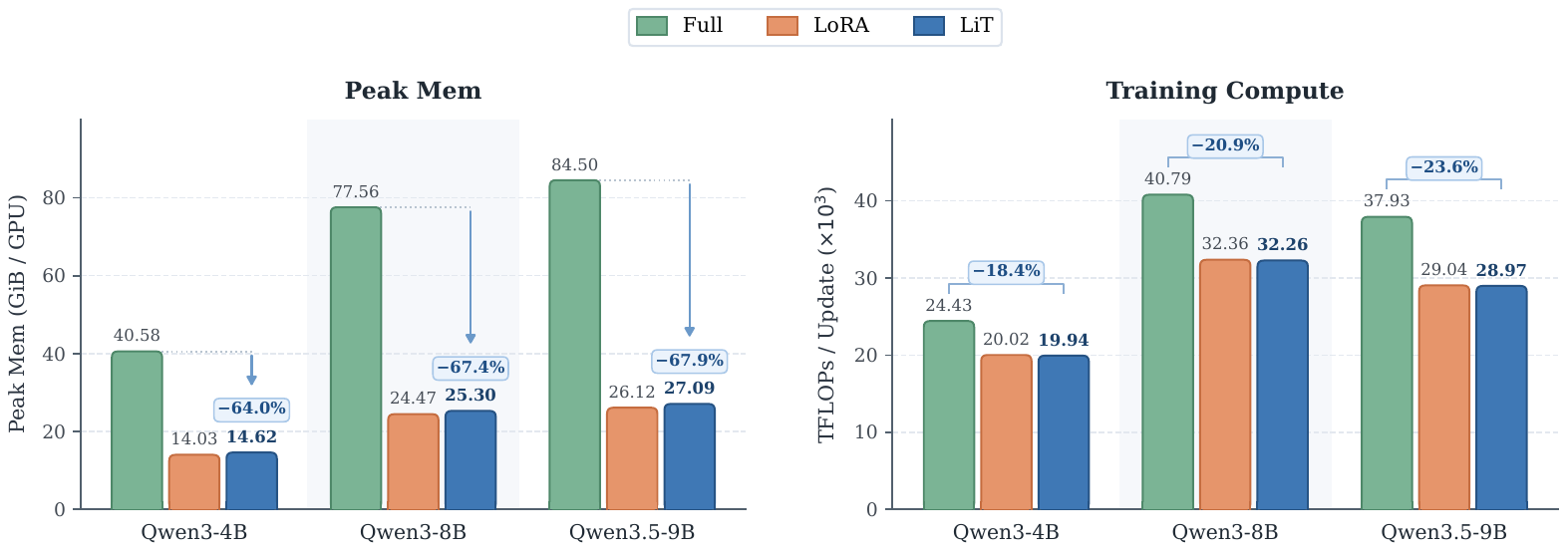}
    \caption{
    Training efficiency comparison across three model scales. Left: peak allocated GPU memory~(\textbf{Peak Mem}, in GiB per GPU, taking the maximum across all four GPUs during steady-state training). Right: training compute per optimizer update~(in $10^3$ TFLOPs). Annotated percentages indicate reductions achieved by \method relative to full fine-tuning.
    }
    \label{fig:efficiency_comparison}
\end{figure*}
Figure~\ref{fig:efficiency_comparison} compares per-GPU peak allocated memory~(Peak Mem) and training compute per update across all three model scales. Relative to full fine-tuning, \method substantially lowers memory consumption at every scale, reducing Peak Mem by 64.0\% on Qwen3-4B from 40.58 to 14.62~GiB, by 67.4\% on Qwen3-8B from 77.56 to 25.30~GiB, and by 67.9\% on Qwen3.5-9B from 84.50 to 27.09~GiB. Compared with LoRA, \method achieves closely matched memory efficiency, incurring less than 1.0~GiB of additional memory across all three models, ranging from 0.59 to 0.97~GiB, primarily because \method need to store the coordinate indices corresponding to the selected sparse parameters. For training compute, \method achieves the lowest cost across all three scales, cutting TFLOPs per update relative to full fine-tuning by 18.4\% on Qwen3-4B, 20.9\% on Qwen3-8B, and 23.6\% on Qwen3.5-9B while consistently outperforming LoRA. Because \method computes weight gradients only on the selected sparse coordinates via SDDMM and avoids auxiliary low-rank adapter projections, it delivers LoRA-level memory savings with slightly lower compute overhead.

\begin{figure*}[htbp]
    \centering
    \includegraphics[width=0.78\textwidth]{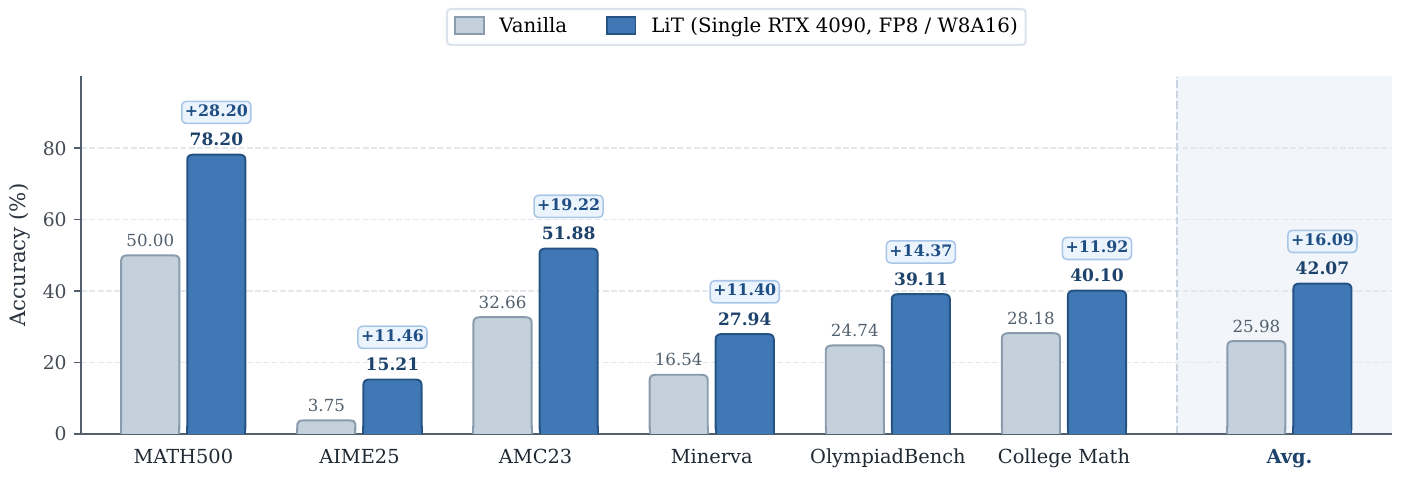}
    \caption{
    Reinforcement learning results on Qwen3-4B using \method with FP8 base quantization and W8A16 computation on a single 24GB RTX 4090D~(maximum rollout and evaluation length of 8,192 tokens). Badges above each \method bar indicate absolute accuracy gains over Vanilla across all six mathematical reasoning benchmarks and their unweighted average~(Avg.).
    }
    \label{fig:4090d_rl_performance}
\end{figure*}

\paragraph{Single-GPU Training Feasibility.}
We additionally evaluate \method on Qwen3-4B with GRPO using a single 24GB RTX 4090D GPU under a separate memory-efficient configuration with FP8 base quantization, W8A16 computation, a trainable parameter ratio of 0.05\% over the target weights~(approximately 1.82M trainable parameters), a learning rate of $2 \times 10^{-4}$, and a maximum rollout length of 8,192 tokens. As shown in Figure~\ref{fig:4090d_rl_performance}, \method raises average accuracy from 25.98 to 42.07, a gain of 16.09 points, with consistent improvements across all six mathematical reasoning benchmarks. In particular, MATH500 rises from 50.00 to 78.20, while AIME25 and AMC23 improve from 3.75 to 15.21 and from 32.66 to 51.88, respectively. These results demonstrate that \method enables substantial RL reasoning gains on a multi-billion-parameter model using a single consumer-grade GPU, providing a practical path toward local model training and inference.

\section{Empirical Study}
\label{sec:empirical_study}

To understand why \method{} exhibits less forgetting and cross-task interference than LoRA, we trace its updates on TRACE using Qwen3-8B from weight-space geometry to layer-wise activations and model outputs. Unless otherwise specified, \method{} and rank-1 LoRA~($\gamma=2$) share a budget of 2.7M trainable parameters, with full fine-tuning~(Full) included as a reference for sequential training. Weight- and activation-space diagnostics cover the 252 targeted attention and MLP projection matrices across all 36 layers, whereas output diagnostics evaluate the complete models.

\begin{figure}[htbp]
    \centering
    \includegraphics[width=0.8\linewidth]{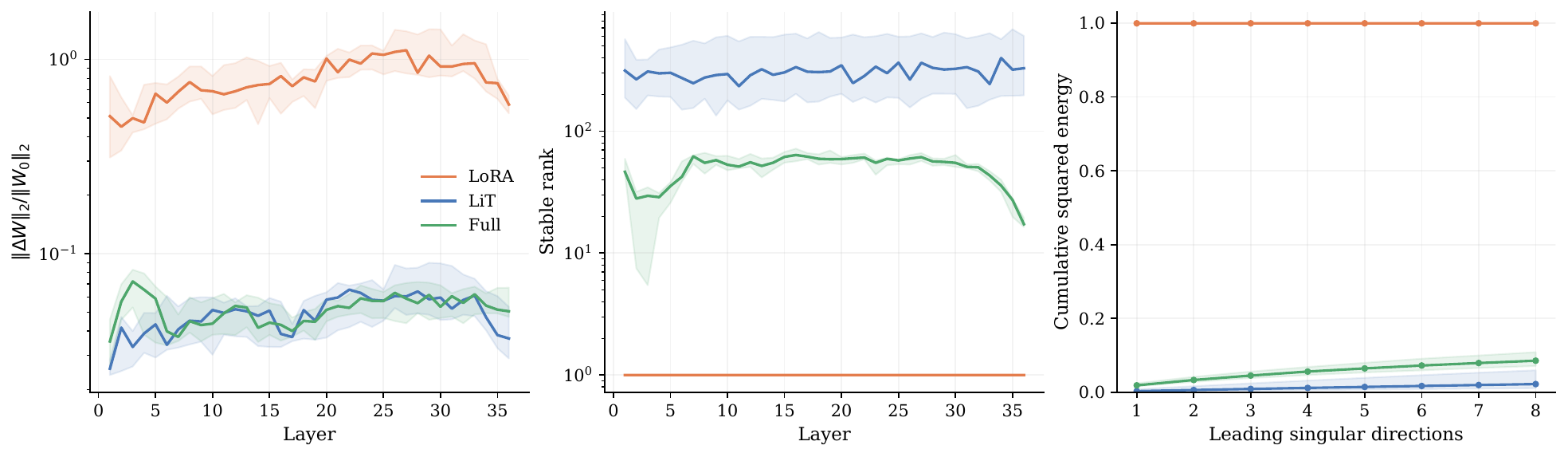}
    \caption{
    Spectral structure of updates after sequential training. Left: relative spectral norm. Middle: stable rank. Curves in these two panels show the median over seven projections per layer, with interquartile shading. Right: cumulative squared energy in the leading $k$ singular directions, normalized by total update energy, with median and interquartile range over 252 matrices.
    }
    \label{fig:mechanism_spectra}
\end{figure}

\paragraph{Update spectra.}
We first inspect the final sequential updates $\Delta W_\ell = W_{\mathrm{final},\ell} - W_{0,\ell}$ across the 252 targeted projections~(Figure~\ref{fig:mechanism_spectra}). Measured by the aggregate Frobenius norm $\big(\sum_{\ell=1}^{252}\|\Delta W_\ell\|_F^2\big)^{1/2}$, \method{} and LoRA induce comparable total weight changes of 139.13 and 145.74, respectively, both larger than Full at 60.68 on the same projections. Despite similar overall update magnitudes, their spectral distributions differ sharply. Because rank-1 LoRA is structurally confined to a single update direction, its median stable rank $\|\Delta W\|_F^2/\|\Delta W\|_2^2$ stays at 1.00 and its median relative spectral norm $\|\Delta W\|_2/\|W_0\|_2$ reaches 0.7883. In contrast, \method{} disperses a comparable update norm across many directions, yielding a median stable rank of 305.35 and placing only 2.24\% of squared update energy in the leading eight singular directions. This dispersion keeps its median relative spectral norm at 0.0485---over $16\times$ smaller than LoRA and close to Full at 0.0509, which has a median stable rank of 52.88 and 8.57\% energy in the top eight directions.

\begin{figure}[htbp]
    \centering
    \includegraphics[width=\linewidth]{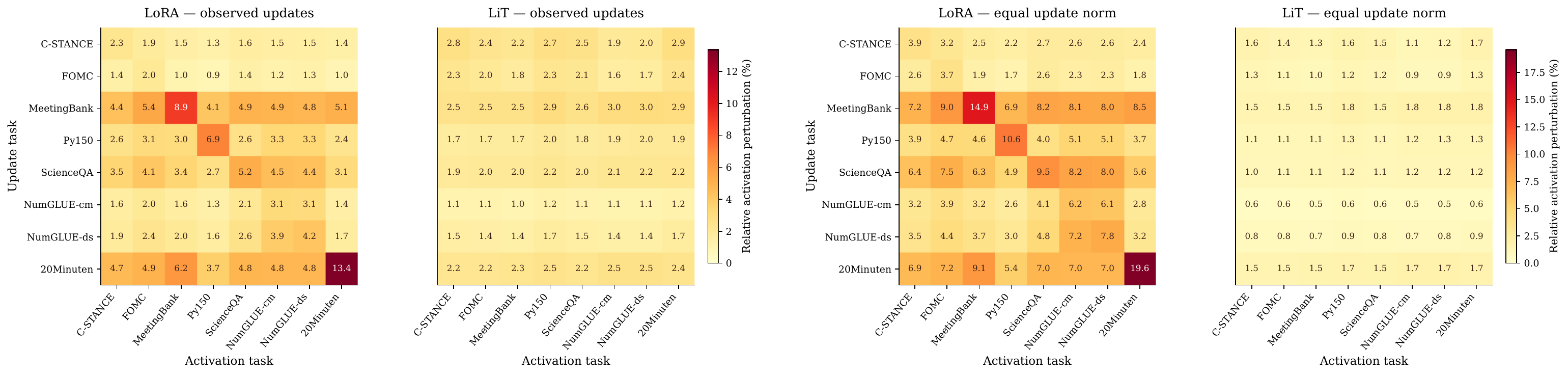}
    \caption{
    Cross-task activation perturbations from independent updates. Rows identify the source task of an update, and columns identify the task supplying base-model activations. Values are $100R_{ij}$, computed from the final prompt token across eight tasks; off-diagonal entries measure cross-task perturbations. Left two panels: observed updates. Right two panels: updates rescaled to equal aggregate Frobenius norm within each source task.
    }
    \label{fig:mechanism_interference}
\end{figure}

\paragraph{Cross-task activation perturbations.}
Moving from weight space to hidden representations, we test how an update learned for task~$i$ perturbs base-model activations from task~$j$ using the eight independently trained models. Let $\Delta W_{i,\ell}$ denote the update for task~$i$ at projection~$\ell$, and let $X_{j,\ell}$ stack the base model's projection inputs at the final prompt token across all test examples of task~$j$~(7,794 examples in total). We measure the relative activation perturbation
\begin{equation}
    R_{ij}=
    \left(
    \frac{
        \sum_{\ell=1}^{252}
        \|X_{j,\ell}\Delta W_{i,\ell}^{\top}\|_F^2
    }{
        \sum_{\ell=1}^{252}
        \|X_{j,\ell}W_{0,\ell}^{\top}\|_F^2
    }
    \right)^{1/2}.
    \label{eq:mechanism_perturbation}
\end{equation}
As shown in Figure~\ref{fig:mechanism_interference}, the mean off-diagonal perturbation across all 56 cross-task pairs is 2.006\% for \method{} versus 2.880\% for LoRA. To rule out differences in update magnitude, we rescale each method's update for a given source task to the geometric mean of the two methods' aggregate Frobenius norms on that task. Under matched update norms, \method{} reduces the mean cross-task perturbation to 1.182\% compared with 4.868\% for LoRA, a 75.7\% reduction, with \method{} lower in all 56 cross-task pairs. Thus, the lower local interference of \method{} reflects the structure of its updates rather than smaller overall update norms.

\begin{figure}[htbp]
    \centering
    \includegraphics[width=0.8\linewidth]{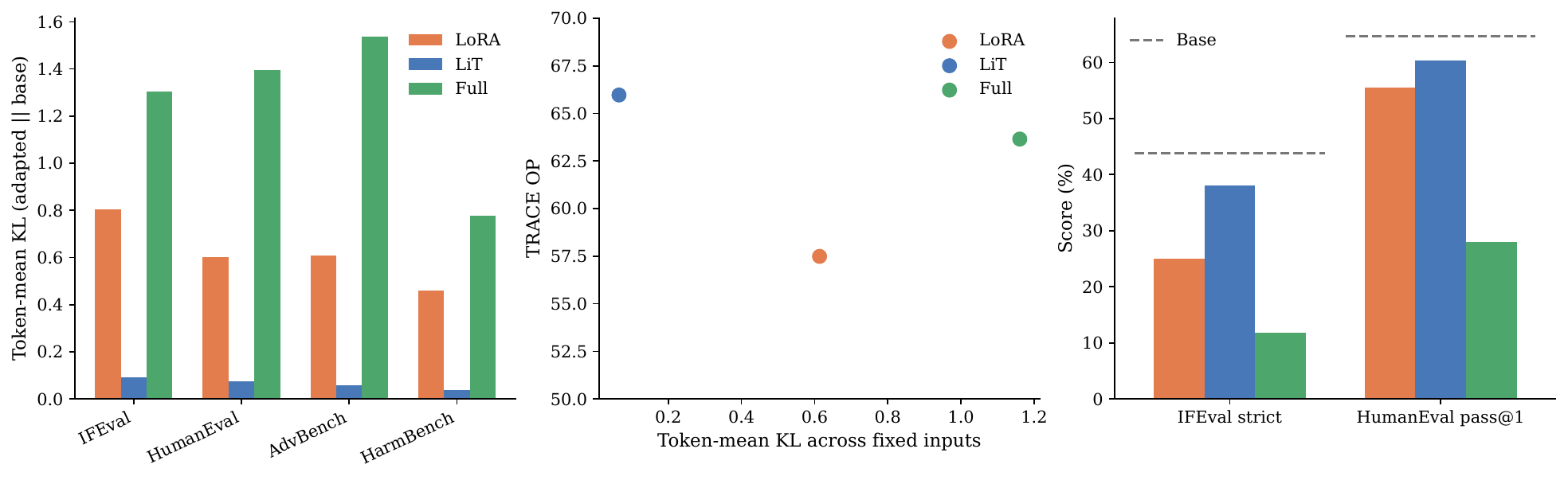}
    \caption{
    Output distribution drift and capabilities after sequential training. Left: token-averaged $D_{\mathrm{KL}}(P_m\|P_0)$ on fixed input prefixes, in nats. Middle: TRACE overall performance against KL pooled across all evaluated token positions. Right: IFEval prompt-level strict accuracy and HumanEval greedy pass@1; dashed lines indicate base-model performance. KL on safety-benchmark prefixes measures distribution drift, not unsafe response or refusal rates.
    }
    \label{fig:mechanism_output}
\end{figure}

\paragraph{Output drift and capability preservation.}
To examine whether these weight- and activation-level differences translate into preserved model behavior, we measure the token-averaged KL divergence $D_{\mathrm{KL}}(P_m\|P_0)$ between each final sequential model and the base model over identical input prefixes~(Figure~\ref{fig:mechanism_output}). Across 95,249 non-final prefix tokens drawn from IFEval, HumanEval, AdvBench, and HarmBench, \method{} incurs a mean KL of 0.0654---nearly an order of magnitude smaller than LoRA at 0.6134 and far below Full at 1.1605---while achieving the highest TRACE OP at 65.97, compared with 57.49 for LoRA and 63.65 for Full. This smaller distribution drift aligns closely with general capability retention: relative to base-model scores of 43.81 on IFEval and 64.63 on HumanEval, \method{} retains 38.08 and 60.37, substantially outperforming both LoRA at 24.95 and 55.49 and Full at 11.83 and 28.05. Notably, Full has nearly the same median relative spectral norm as \method{} yet exhibits the largest output drift and steepest general-capability drop, indicating that small spectral norm alone does not guarantee functional preservation.

\paragraph{Budget scaling and update geometry.}
Finally, we track how these geometric and output metrics vary with the trainable parameter ratio $p$, under the budget-scaled learning-rate schedule in Appendix~\ref{app:lr_scaling}. The geometric trajectory mirrors the non-monotonic performance trend in Figure~\ref{fig:trace_parameter_budget}. At extreme sparsity~($p = 10^{-7}$), updates concentrate on a handful of large entries, producing a median stable rank of 1.23 and a relative spectral norm of 1.67. Across the intermediate sparsity window, update energy disperses rapidly: at $p = 10^{-2}$, stable rank peaks at 902.04 and relative spectral norm reaches a minimum of 0.0203 alongside the peak TRACE OP of 66.55, whereas fixed-prefix KL attains its lowest value of 0.0654 at the main budget $p \approx 3.33\times 10^{-4}$. Increasing density further to $p = 0.5$ reverses these trends, pulling stable rank down to 77.43, driving KL up to 0.9499, and lowering OP to 62.34.

\begin{figure}[htbp]
    \centering
    \includegraphics[width=0.8\linewidth]{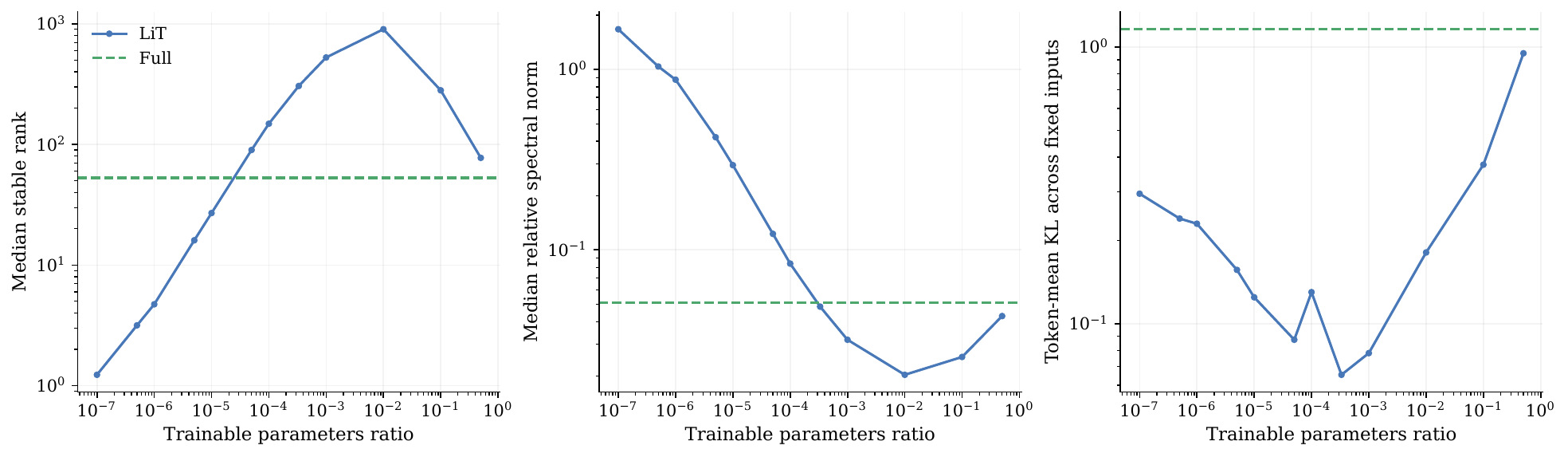}
    \caption{
    Update geometry and output drift across \method{} parameter budgets. Left: median stable rank over the target projections. Middle: median relative spectral norm. Right: token-averaged KL on fixed input prefixes. Both axes use logarithmic scales. The trainable parameter ratio is relative to the full-model parameter count; $p=10^{-2}$ denotes 1\%. Green dashed lines show the single Full reference configuration. Learning rates follow the inverse-square-root budget rule in Appendix~\ref{app:lr_scaling}.
    }
    \label{fig:mechanism_budget}
\end{figure}

\section{Related Work}
\label{sec:related_work}

\paragraph{Efficient Training.}
Parameter-efficient training aims to reduce the computational and memory costs of model adaptation. LoRA series~\citep{hu2021lora, zhang2023adalora, liu2024dora, meng2024pissa} achieve this through low-rank weight updates. Diff Pruning~\citep{guo-etal-2021-parameter} learns sparse task-specific weight updates through differentiable pruning, while FISH Mask~\citep{sung2021trainingneuralnetworksfixed} selects trainable parameters based on Fisher information. More closely related to our work, Random Masking~\citep{xu2024randommasking}, Sparse Random Parameter Adaptation~\citep{rios2025sparsity}, and SHiRA~\citep{bhardwaj2024shira} demonstrate that random parameter selection can achieve performance comparable to LoRA. Our work goes further to demonstrate that uniformly random sparse weight updates can also support effective reinforcement learning. Moreover, we extend this paradigm to continual learning, showing reduced catastrophic forgetting during sequential training and better preservation of multi-task capabilities when independently learned updates are merged.

\paragraph{Continual Learning.}
Continual learning aims to enable models to acquire new knowledge and capabilities while retaining previously learned ones, balancing plasticity and stability. Traditional approaches address this challenge through regularization, replay, and parameter isolation~\citep{kirkpatrick2017overcoming, lopezpaz2017gem, mallya2018packnet}. More recently, parameter-efficient methods for continual learning have been explored. $\delta$-mem~\citep{lei2026delta} compresses historical information into a dynamic matrix. Progressive Prompts~\citep{razdaibiedina2023progressive} incrementally learns task-specific prompts while freezing the backbone. O-LoRA~\citep{wang2023olora} learns task-specific updates in mutually orthogonal low-rank subspaces, while InfLoRA~\citep{liang2024inflora} constructs interference-free update. Beyond sequential learning, model merging provides another approach to integrating independently acquired knowledge. Task Arithmetic~\citep{ilharco2023taskarithmetic} combines task-specific weight updates, while TIES-Merging~\citep{yadav2023ties} addresses interference among merged parameters. Unlike approaches that explicitly constrain update subspaces or isolate task-specific parameters, our work shows that uniformly random sparse weight updates can effectively support both sequential learning and independent knowledge integration. In addition, we examine their ability to maintain general capabilities and safety-related behaviors during continued adaptation.

\section{Conclusion}
\label{sec:conclusion}

In this work, we introduced \method, a simple yet effective training approach that directly optimized a uniformly random subset of original weight coordinates without relying on low-rank parameterization or importance-based parameter selection. By learning sparse additive updates in the original weight space and computing gradients only for selected coordinates, \method enabled parameter-efficient adaptation while reducing the associated optimization costs. Across both sequential continual learning and multi-task model merging, \method substantially reduced catastrophic forgetting and cross-task interference while preserving general capabilities, factual knowledge, and safety boundaries far better than LoRA. Beyond supervised adaptation, the same sparse parameterization transferred directly to reinforcement learning, outperforming both LoRA and full fine-tuning across all evaluated settings and making 4B-scale RL feasible on a single consumer-grade GPU. By effectively mitigating catastrophic forgetting as well as the degradation of safety alignment and general capabilities during task training, \method opens a practical new path toward training personal continual learning intelligence.

\section{Future Work}
\label{sec:future_work}

While \method{} provides an effective basis for continual adaptation, two directions merit further investigation.

\paragraph{Simple and Effective Parameter Merging.}
Our multi-task integration experiments combine independently trained updates via direct unweighted summation. Because each LiT update modifies only a tiny fraction of coordinates, independently sampled supports across tasks are nearly disjoint, yielding minimal direct coordinate collision even as the number of tasks grows. Consequently, rather than relying on complex conflict-resolution rules for overlapping dense updates, future work can focus on lightweight calibration of cumulative activation and spectral drift, such as scale-aware weighting or simple coordinate consolidation, to further reduce interference as the number of tasks or user-specific updates scales.

\paragraph{Toward Self-Modifying Models.}
\method{} demonstrates that acquiring new capabilities and refining model behavior require updating only a tiny fraction of native weights directly in the original parameter space, without auxiliary adapter structures. By substantially reducing the dimensionality and structural overhead of weight edits, this native sparse-update paradigm may inspire self-modifying models that directly generate and apply compact sparse weight updates to continuously internalize new experiences.

\bibliographystyle{plainnat}
\bibliography{references}

\newpage
\appendix

\section{Continual Learning Settings}
\label{app:cl-settings}
We use AdamW with $\beta_1=0.9$, $\beta_2=0.999$, zero weight decay, global batch size is set to 128. The learning rate remains constant without warmup, and gradients are clipped at a global norm of 1.0. Training uses BF16, with FP32 gradient accumulation and communication, and dropout is disabled. We train on all eight tasks in the standard order (\textit{C-STANCE}, \textit{FOMC}, \textit{MeetingBank}, \textit{Py150}, \textit{ScienceQA}, \textit{NumGLUE-cm}, \textit{NumGLUE-ds}, \textit{20Minuten}) using $5{,}000$ training examples per task and the official per-task epoch schedule $(5, 3, 7, 5, 3, 5, 5, 7)$. Prompts are capped at $1{,}024$ tokens~(retaining the first $128$ and last $896$ tokens when truncated to preserve both instructions and answer cues), answers are capped at $512$ tokens, and the cross-entropy loss is computed over answer tokens only. Let $S_{t,i}$ denote the evaluation score on task $i$ after sequentially training through task $t$ ($1 \le i \le t \le T$, with $T=8$ on TRACE). Following the TRACE benchmark convention, we define the final overall performance~(OP) and backward transfer~(BWT) as
\begin{equation}
\mathrm{OP} = \frac{1}{T}\sum_{i=1}^{T} S_{T,i}, \qquad
\mathrm{BWT} = \frac{1}{T}\sum_{i=1}^{T-1}\left(S_{T,i} - S_{i,i}\right).
\end{equation}

\section{Reinforcement Learning Settings}
\label{app:rl-settings}

\paragraph{Training Setup.}
We conduct RL on 29,427 problems from OpenThoughts Math~\citep{guha2026openthoughts} for 100 optimizer steps, with a maximum generation length of 16,384 tokens and a sampling temperature of 1.0. Each update uses a global batch of 256 responses~(32 prompts with 8 responses per prompt) and performs one gradient step per rollout. Correctness is scored by a rule-based binary verifier that extracts answers from \verb|\boxed{...}| and checks equivalence via normalization and symbolic matching (1 if correct, 0 otherwise), with no format reward or learned reward model. Both algorithms use group-relative advantage normalization with zero KL penalty and zero entropy regularization. Specifically, our GRPO implementation uses the 0/1 reward directly with a REINFORCE objective, score centering~\citep{marek2026score} ($\text{top-}k=128$), response-level loss averaging, and no PPO clipping. For DAPO, correctness is mapped to $\pm 1$ with a soft overlong penalty that linearly increases from 0 at 13,108 tokens to $-1$ at the 16,384-token limit (buffer $B = \lfloor 0.2 L_{\max} \rfloor = 3,276$); DAPO further employs PPO clipped loss ($\epsilon_{\mathrm{low}}=0.2$, $\epsilon_{\mathrm{high}}=0.28$), token-level loss averaging, and dynamic sampling (filtering out prompt groups with identical rewards and resampling until 32 valid groups are collected). All runs use AdamW with zero weight decay, gradient clipping at 1.0, a constant learning rate schedule (no warmup or decay), and random seed 42.

\paragraph{Evaluation.}
All evaluations are performed using sampling decoding with a maximum of 16,384 new tokens (excluding prompts), temperature 1.0, $\text{top-}p=0.95$, no top-$k$ truncation, and random seed 42. We sample 1 response per problem for MATH500, Minerva, OlympiadBench, and College Math, and 16 responses per problem for AIME 2025 and AMC23, reporting mean accuracy across the 16 samples (without majority voting or pass@16). We report the unweighted average across all six benchmarks.

\section{Reference-Anchored Learning Rate Adoption}
\label{app:lr_scaling}
We found that the optimal learning rate for \method varies across parameters. Conducting an exhaustive learning rate sweep for every parameter budget is computationally prohibitive. Instead, motivated by the update dynamics of AdamW, we adopt the following: anchored at a well-performing learning rate from our main experiments ($lr = 10^{-3}$), to estimate an effective learning rate for other parameter budgets.

\paragraph{Reference Configuration.}
We anchor the scaling rule to the Qwen3-8B \textsc{\method} configuration at the rank-1-matched parameter budget:
\begin{equation}
    K_0 = 2{,}727{,}936, \qquad lr = 10^{-3},
\end{equation}
where $K_0$ is the number of trainable parameters and $lr$ is the reference learning rate.

\paragraph{Square-Root Scaling under AdamW.}
With weight decay set to $0$, the single-step AdamW update for each trainable coordinate $i \in \{1, \dots, K\}$ is
\begin{equation}
    \Delta \theta_i = -lr \cdot \, u_i, \qquad u_i = \frac{\hat{m}_i}{\sqrt{\hat{v}_i} + \epsilon}.
\end{equation}
The $\ell_2$ norm of the single-step parameter update is therefore
\begin{equation}
    \|\Delta \theta\|_2 = lr \cdot \sqrt{\sum_{i=1}^{K} u_i^2} = lr \cdot \sqrt{K} \, \mathrm{RMS}(u).
\end{equation}
Assuming $\mathrm{RMS}(u)$ remains roughly comparable across parameter budgets, maintaining a similar single-step update magnitude yields the scaling rule:
\begin{equation}
    \label{eq:lr_scaling}
    lr(K) = lr \cdot \sqrt{\frac{K_0}{K}} = 10^{-3}\sqrt{\frac{2{,}727{,}936}{K}}.
\end{equation}
Because \textsc{\method} directly updates selected weight coordinates, $\|\Delta \theta\|_2$ corresponds directly to the Frobenius norm of the effective weight update.

\paragraph{Budget Instantiation.}
For Qwen3-8B, the full model contains $N_{\text{total}} = 8,190,735,360$ parameters. Given a target trainable ratio $p\%$ relative to $N_{\text{total}}$, we allocate $K \approx N_{\text{total}} p/100$ coordinates across the targeted projections and then compute $lr(K)$ via Eq.~\eqref{eq:lr_scaling}. The resulting parameter counts and learning rates across all evaluated budgets are listed in Table~\ref{tab:parameter_budget}.

\begin{table}[t]
    \centering
    \caption{
    Trainable parameter budgets and learning rates for the TRACE parameter-scaling experiment on Qwen3-8B.
    }
    \label{tab:parameter_budget}

    \setlength{\tabcolsep}{4pt}
    \renewcommand{\arraystretch}{1.12}

    \resizebox{0.6\textwidth}{!}{%
    \begin{tabular}{l r r c}
        \toprule
        \textbf{Method}
        & \textbf{Ratio (\%)}
        & \textbf{\# Trainable Params}
        & \textbf{Learning Rate} \\
        \midrule

        LiT & 0.00001 & 819 & $5.771\times10^{-2}$ \\
        LiT & 0.00005 & 4,095 & $2.581\times10^{-2}$ \\
        LiT & 0.0001 & 8,191 & $1.825\times10^{-2}$ \\
        LiT & 0.0005 & 40,954 & $8.161\times10^{-3}$ \\
        LiT & 0.001 & 81,907 & $5.771\times10^{-3}$ \\
        LiT & 0.005 & 409,537 & $2.581\times10^{-3}$ \\
        LiT & 0.01 & 819,074 & $1.825\times10^{-3}$ \\
        \rowcolor{sqaDeltaBg}
        \textbf{LiT (rank-1 budget)} & \textbf{0.0333} & \textbf{2,727,936} & $\mathbf{1\times10^{-3}}$ \\
        LiT & 0.1 & 8,190,735 & $5.771\times10^{-4}$ \\
        LiT & 1 & 81,907,354 & $1.825\times10^{-4}$ \\
        LiT & 10 & 819,073,536 & $5.771\times10^{-5}$ \\
        LiT & 50 & 4,095,367,680 & $2.581\times10^{-5}$ \\
        \midrule

        Full & 100 & 8,190,735,360 & $1.825\times10^{-5}$ \\
        \bottomrule
    \end{tabular}%
    }
\end{table}

\section{RL Training Dynamics}
\begin{figure*}[htbp]
    \centering
    \includegraphics[width=0.8\textwidth]{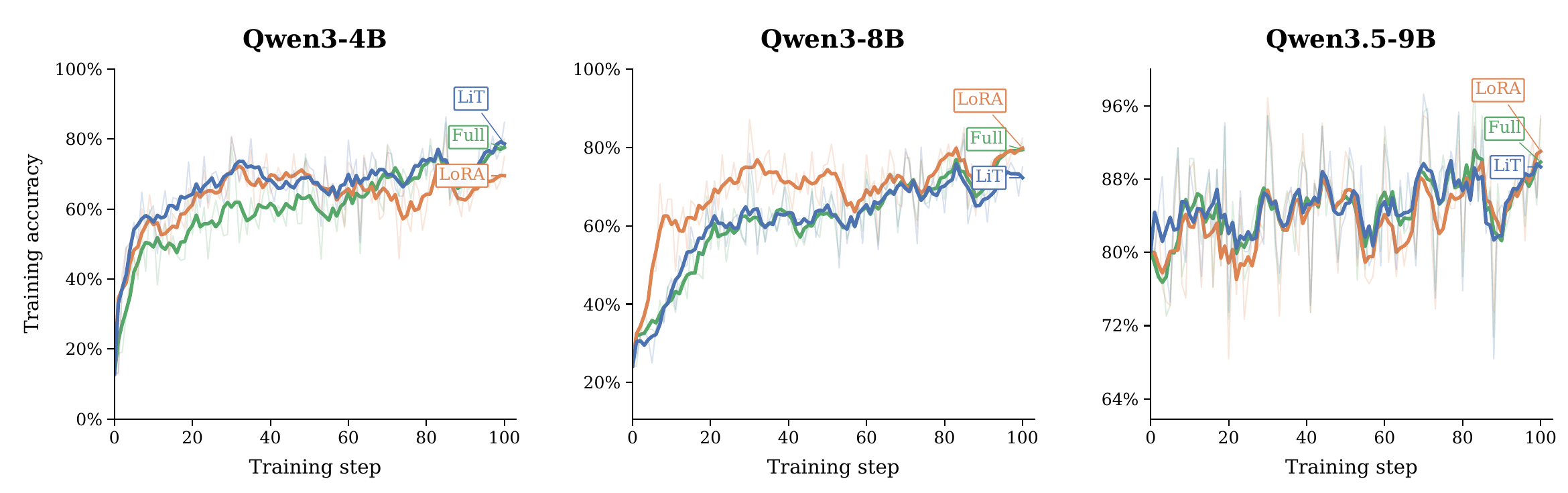}
    \caption{
    Training accuracy during GRPO training across different model scales.
    We compare \method, LoRA, and full fine-tuning on Qwen3-4B,
    Qwen3-8B, and Qwen3.5-9B. The horizontal axis represents
    training steps, and the vertical axis represents training accuracy.
    }
    \label{fig:grpo_training_accuracy}
    \vspace{-3mm}
\end{figure*}

\paragraph{Training Accuracy Dynamics.}
Figure~\ref{fig:grpo_training_accuracy} compares the evolution of training accuracy during GRPO training across three model scales. On Qwen3-4B, \method achieves training accuracy comparable to full fine-tuning and consistently higher than LoRA toward the end of training. On Qwen3-8B, its training accuracy remains competitive throughout training. On Qwen3.5-9B, \method shows a steady increase in training accuracy and eventually approaches the performance of full fine-tuning. 

\begin{figure*}[htbp]
    \centering
    \includegraphics[width=0.8\textwidth]{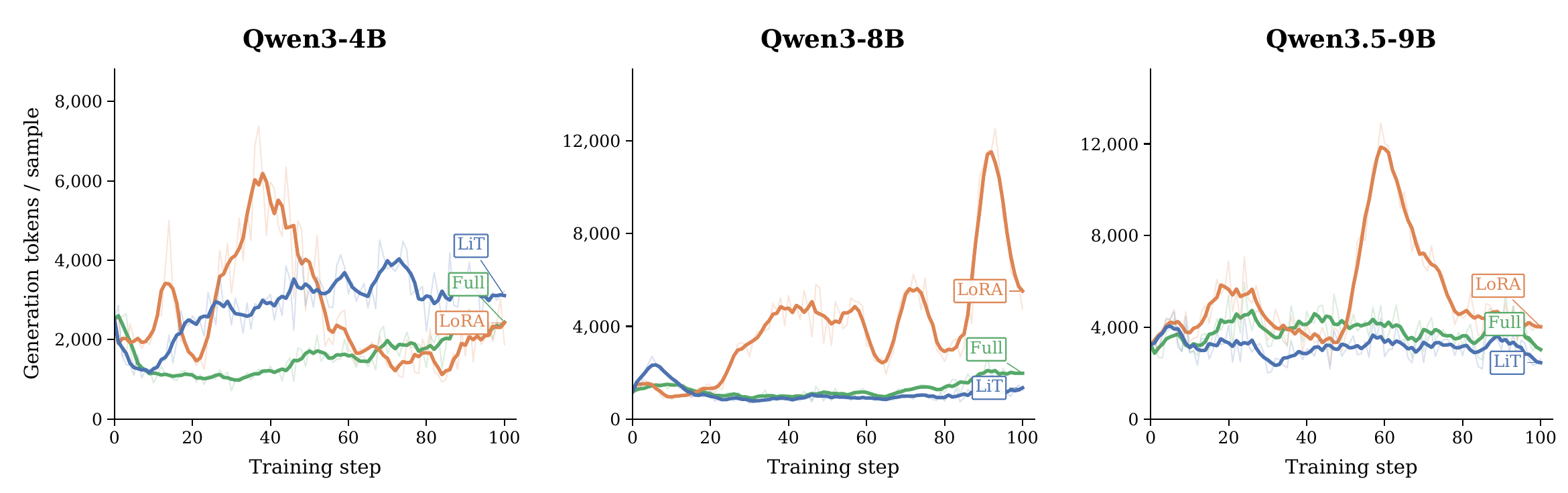}
    \caption{
    Average generation length per sample during GRPO training across
    different model scales. We compare \method, LoRA, and full
    fine-tuning on Qwen3-4B, Qwen3-8B, and Qwen3.5-9B.
    The curves show the evolution of generated tokens per sample
    throughout training.
    }
    \label{fig:grpo_generation_tokens}
    \vspace{-3mm}
\end{figure*}

\paragraph{Generation Length Dynamics.}
Figure~\ref{fig:grpo_generation_tokens} illustrates the evolution of average generation length per sample throughout GRPO training. Across the three model scales, LoRA exhibits pronounced fluctuations in generation length, with particularly large spikes on Qwen3-8B and Qwen3.5-9B, reaching approximately 12,000 tokens per sample. In comparison, \method generally maintains a more moderate generation length and avoids these large spikes like full fine-tuning, particularly on the larger models. On Qwen3-8B, for example, the generation length of \method remains around 1,000--2,000 tokens toward the end of training, whereas LoRA exhibits a substantial increase. These observations suggest that \method can achieve competitive reasoning performance without the large generation-length fluctuations observed with LoRA.

\begin{figure*}[htbp]
    \centering
    \includegraphics[width=0.8\textwidth]{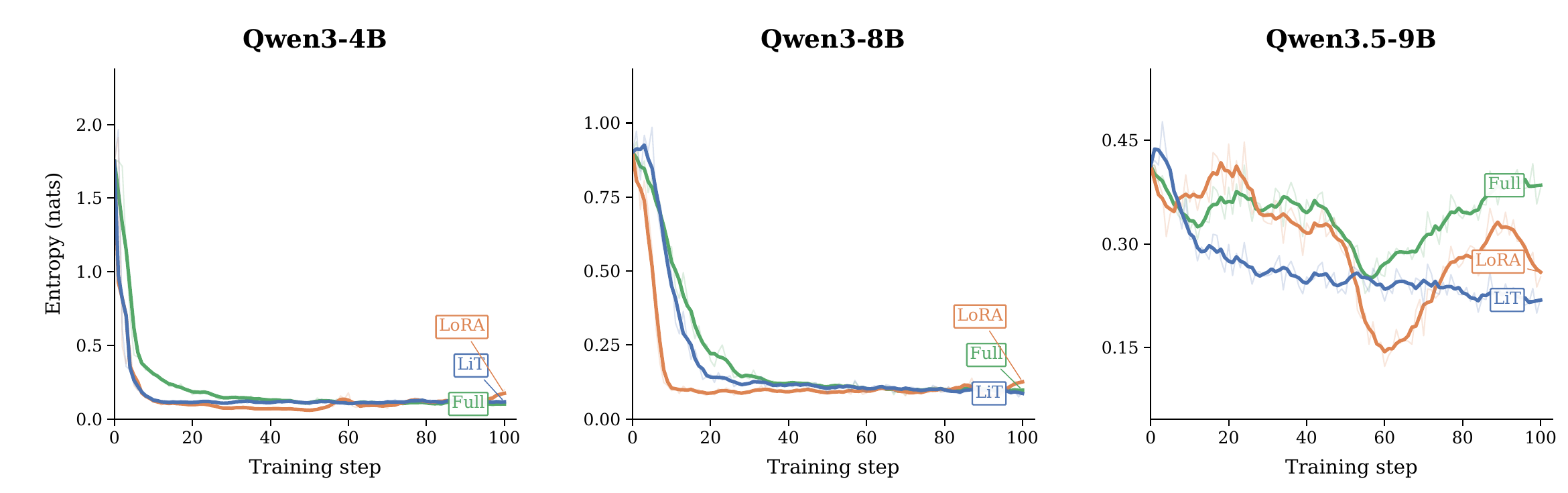}
    \caption{
    Policy entropy during GRPO training across different model scales. We compare \method, LoRA, and full fine-tuning on Qwen3-4B, Qwen3-8B, and Qwen3.5-9B. Entropy is measured in nats, with the horizontal axis representing training steps.
    }
    \label{fig:grpo_entropy}
    \vspace{-3mm}
\end{figure*}

\paragraph{Policy Entropy Dynamics.}
Figure~\ref{fig:grpo_entropy} compares policy entropy
throughout GRPO training across different model scales.
On Qwen3-4B and Qwen3-8B, all three methods exhibit a rapid decrease in entropy during early training, followed by convergence to relatively low entropy levels. On Qwen3.5-9B, their trajectories become more distinct: the entropy of \method decreases gradually and remains relatively stable during later training, while LoRA exhibits a more pronounced mid-training decrease followed by a rebound. Full fine-tuning also shows an increase in entropy toward the end of training.

\section{Results Across Random Seeds}
\begin{table*}[htbp]
    \centering
    \caption{
    Seven-seed evaluation of \method{} on Qwen3-8B under sequential learning on TRACE.
    All eight tasks are learned sequentially using a learning rate of $1\times10^{-3}$.
    We report task-wise performance, overall performance~(OP), and backward transfer~(BWT) across seven
    random seeds (42--48).
    Higher values indicate better performance.
    }
    \label{tab:multi_seed}

    \setlength{\tabcolsep}{5.5pt}
    \renewcommand{\arraystretch}{1.18}

    \resizebox{0.7\textwidth}{!}{%
    \begin{tabular}{
        l @{\hspace{8pt}}
        *{7}{c}
        @{\hspace{10pt}} c
    }
        \toprule

        \multirow{2}{*}{\textbf{Task}}
        & \multicolumn{7}{c}{\textbf{Random Seed}}
        & \multirow{2}{*}{\textbf{Mean $\pm$ Std}} \\
        \cmidrule(lr){2-8}

        & \textbf{42}
        & \textbf{43}
        & \textbf{44}
        & \textbf{45}
        & \textbf{46}
        & \textbf{47}
        & \textbf{48}
        & \\

        \midrule

        C-STANCE
        & 55.45 & 55.65 & 56.85 & 57.30
        & 55.85 & 53.30 & 56.15
        & \cellcolor{sqaDeltaBg}55.79 $\pm$ 1.28 \\

        FOMC
        & 68.55 & 66.73 & 66.13 & 65.52
        & 68.35 & 68.15 & 70.16
        & \cellcolor{sqaDeltaBg}67.66 $\pm$ 1.61 \\

        MeetingBank
        & 56.55 & 57.16 & 57.87 & 57.17
        & 57.24 & 55.28 & 58.74
        & \cellcolor{sqaDeltaBg}57.14 $\pm$ 1.07 \\

        Py150
        & 67.06 & 67.06 & 67.53 & 67.55
        & 67.17 & 67.32 & 67.67
        & \cellcolor{sqaDeltaBg}67.34 $\pm$ 0.25 \\

        ScienceQA
        & 91.00 & 90.20 & 90.40 & 91.15
        & 90.45 & 90.10 & 90.70
        & \cellcolor{sqaDeltaBg}90.57 $\pm$ 0.40 \\

        NumGLUE-cm
        & 77.78 & 80.25 & 72.84 & 75.31
        & 76.54 & 74.07 & 74.07
        & \cellcolor{sqaDeltaBg}75.84 $\pm$ 2.56 \\

        NumGLUE-ds
        & 70.46 & 69.23 & 70.77 & 69.54
        & 72.00 & 69.23 & 69.23
        & \cellcolor{sqaDeltaBg}70.07 $\pm$ 1.06 \\

        20Minuten
        & 40.90 & 41.83 & 41.18 & 41.41
        & 41.42 & 41.22 & 41.77
        & \cellcolor{sqaDeltaBg}41.39 $\pm$ 0.33 \\

        \midrule

        \textbf{OP $\uparrow$}
        & 65.97
        & 66.01
        & 65.45
        & 65.62
        & 66.13
        & 64.83
        & 66.06
        & \cellcolor{sqaDeltaBg}\textbf{65.72 $\pm$ 0.47} \\

        \textbf{BWT $\uparrow$}
        & $-1.25$
        & $-1.00$
        & $-1.44$
        & $-1.33$
        & $-1.13$
        & $-1.51$
        & $-0.62$
        & \cellcolor{sqaDeltaBg}\textbf{$-$1.18 $\pm$ 0.30} \\

        \bottomrule
    \end{tabular}%
    }
\end{table*}
Table~\ref{tab:multi_seed} evaluates the robustness of \method{} across seven random seeds on TRACE. Holding the training, data-order, and evaluation seeds fixed at 42, we vary only the coordinate-selection seed across seven values (42--48) to sample seven independent random sparse supports. Despite the randomness in parameter selection, \method{} achieves consistent performance across different runs, with an average overall performance of $65.72 \pm 0.47$ and backward transfer of $-1.18 \pm 0.30$. These results suggest that the effectiveness of random sparse weight updates is not highly sensitive to the choice of random seed.

\section{Effect of Coordinate Selection Strategy.}
\begin{table*}[htbp]
    \centering
    \caption{
    Comparison of sparse coordinate selection strategies on sequential TRACE~(Qwen3-8B). All methods share the exact same per-matrix trainable parameter budget~(2.7M parameters) and keep the selected coordinate set fixed across all eight tasks. Wanda and FISH are calibrated on 4,000 training examples~(500 per task across all eight TRACE tasks), whereas \method~(Random) selects coordinates uniformly without any calibration data.
    }
    \label{tab:selection_ablation}

    \setlength{\tabcolsep}{2.5pt}
    \renewcommand{\arraystretch}{1.4}

    \resizebox{0.8\textwidth}{!}{%
    \begin{tabular}{
        l @{\hspace{10pt}}
        cccccccc @{\hspace{12pt}}
        cc
    }
        \toprule
        \textbf{Selection} & \textbf{C-STANCE} & \textbf{FOMC} & \textbf{MeetingBank} & \textbf{Py150} & \textbf{ScienceQA} & \textbf{NumGLUE-cm} & \textbf{NumGLUE-ds} & \textbf{20Minuten} & \textbf{OP} $\uparrow$ & \textbf{BWT} $\uparrow$ \\
        \midrule

        Magnitude
        & 54.85 & 62.50 & 50.03 & 66.15 & 89.75 & 74.07 & 70.15 & 40.72 & 63.53 & $-$3.55 \\

        Wanda
        & 45.80 & 64.72 & 19.94 & 52.24 & 83.70 & 69.14 & 67.08 & \textbf{41.51} & 55.52 & $-$10.64 \\

        FISH
        & 25.40 & 0.00 & 10.12 & 31.91 & 41.15 & 38.27 & 21.23 & 40.43 & 26.06 & $-$38.16 \\

        \midrule
        \rowcolor{lightBlue}
        \textbf{\method~(Random)}
        & \textbf{55.45} & \textbf{68.55} & \textbf{56.55} & \textbf{67.06} & \textbf{91.00} & \textbf{77.78} & \textbf{70.46} & 40.90 & \textbf{65.97} & \textbf{$-$1.25} \\

        \bottomrule
    \end{tabular}%
    }
\end{table*}
Table~\ref{tab:selection_ablation} compares \method~(Random) against Magnitude, Wanda, and FISH Mask on sequential TRACE~(Qwen3-8B) under identical per-matrix parameter budgets. Uniform random selection achieves the highest OP~(65.97) and best BWT~($-$1.25), leading on seven of the eight tasks without any calibration data. Since all four methods reach comparable accuracy on the final task, 20Minuten~(40.43 to 41.51), the degradation is driven by forgetting rather than plasticity: BWT worsens monotonically from $-$1.25~(Random) to $-$3.55~(Magnitude), $-$10.64~(Wanda), and $-$38.16~(FISH), with FOMC collapsing to 0.00 under FISH. This trend reflects a direct conflict between single-task importance criteria and continual learning. Rather than protecting sensitive parameters, Wanda and FISH concentrate all trainable coordinates onto heavy-tailed high-activation channels and high-Fisher bottlenecks; under AdamW's nearly uniform per-coordinate step size, updating these high-leverage shared pathways may amplifies activation perturbations and overwrites earlier tasks. Uniform random selection instead samples from the low-curvature bulk of the weight space, reducing interference while avoiding calibration overhead.

\section{Task-wise Performance Evolution and Forgetting Dynamics}
\label{app:forgetting_dynamics}

To further investigate the forgetting behavior of LoRA and
\method{} during sequential learning, we analyze the
task-wise performance evolution throughout the eight-task
TRACE training sequence.

\begin{figure*}[htbp]
    \centering
    \includegraphics[
        width=0.65\textwidth,
        height=0.78\textheight,
        keepaspectratio
    ]{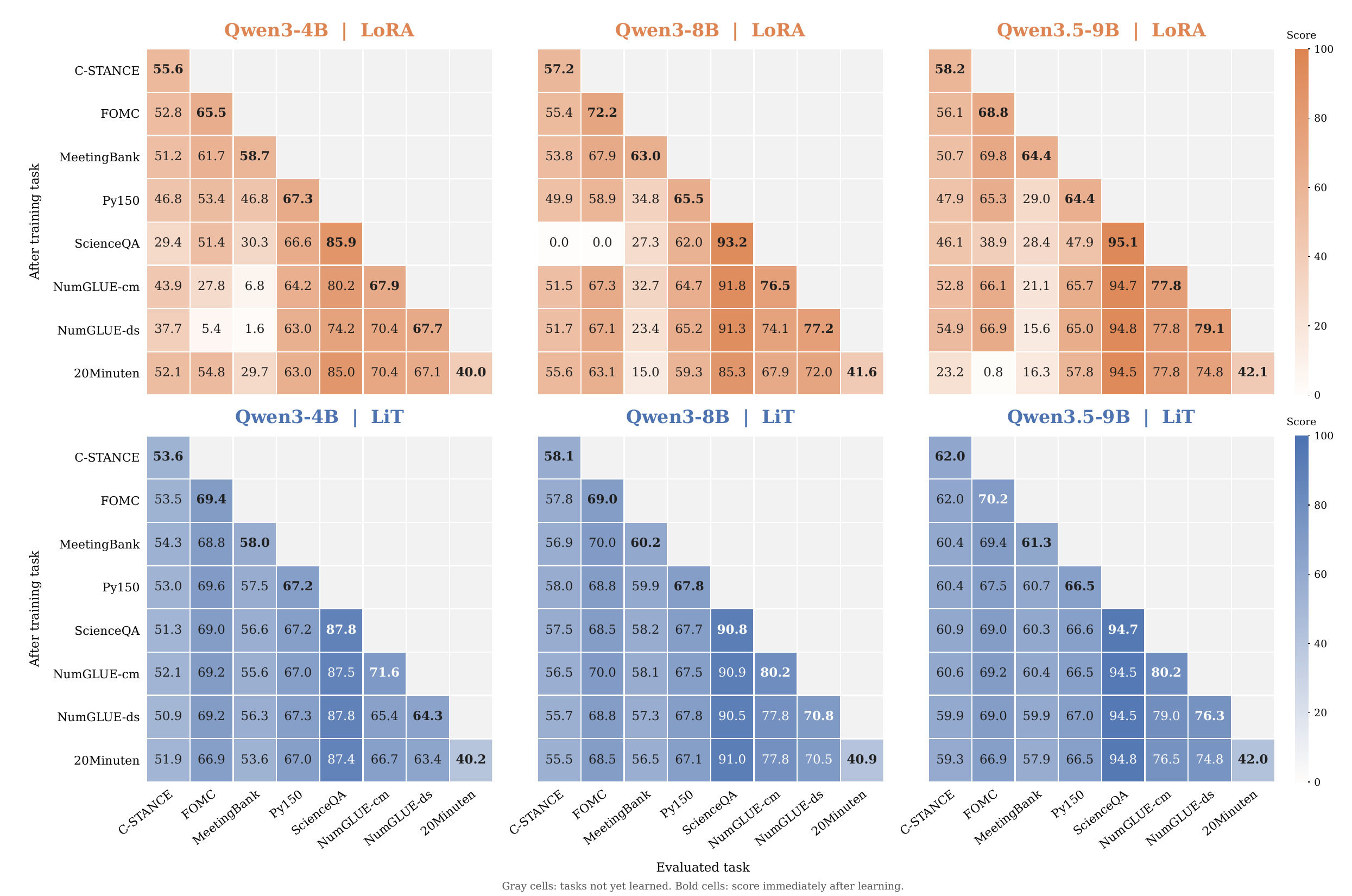}
    \caption{
    Task-wise performance evolution during sequential learning on TRACE across Qwen3-4B, Qwen3-8B, and Qwen3.5-9B. Rows indicate successive training stages, and columns indicate evaluated tasks. The top and bottom rows of panels show LoRA and \method{}, respectively. Bold diagonal entries report scores immediately after learning each task; entries below the diagonal show subsequent performance on that task.
    }
    \label{fig:performance_matrix}
\end{figure*}

\begin{figure}[htbp]
    \centering
    \includegraphics[
        width=0.6\textwidth
    ]{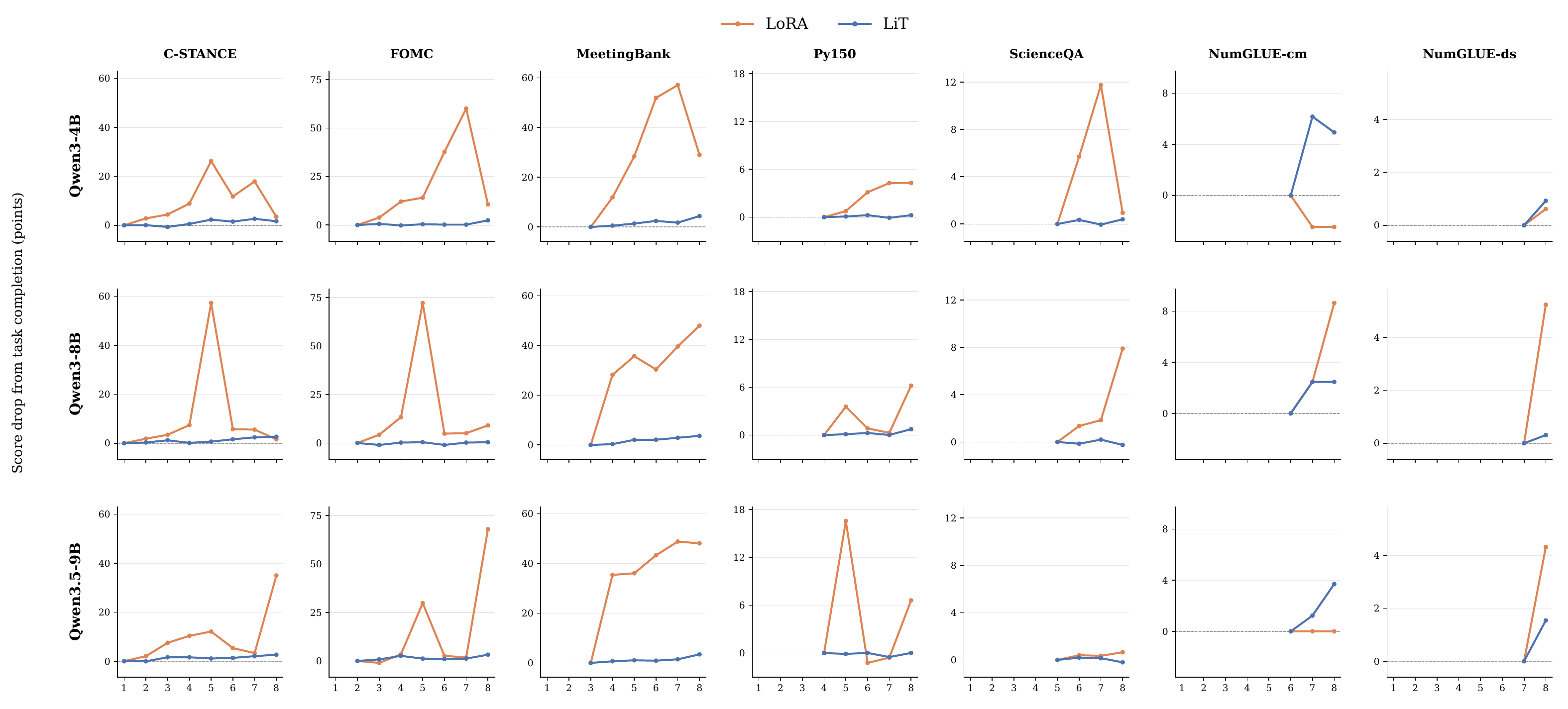}
    \caption{
    Task-wise forgetting trajectories derived from the
    performance scores in Figure~\ref{fig:performance_matrix}. The horizontal axis indicates the sequential training stage, and the vertical axis reports the score drop relative to task completion, $D_{t,i}=S_{i,i}-S_{t,i}$. Positive values indicate forgetting, and negative values indicate improvement. The final task, 20Minuten, is omitted because no subsequent training occurs.
    }
    \label{fig:forgetting_curves}
\end{figure}

We examine how task performance evolves throughout sequential training using two complementary views of the same evaluation results. Figure~\ref{fig:performance_matrix} reports the absolute scores of previously encountered tasks after each training stage, capturing both initial task acquisition and subsequent retention. Figure~\ref{fig:forgetting_curves} expresses these scores as performance drops relative to the point when each task was first learned, enabling a direct comparison of forgetting.

Specifically, let $S_{t,i}$ denote the evaluation score on
task $i$ after training through task $t$.
For $t \geq i$, we define
\begin{equation}
    D_{t,i} = S_{i,i} - S_{t,i},
\end{equation}
where positive values indicate forgetting and negative values
indicate improvement relative to the task-completion baseline.

Across all three models, LoRA exhibits substantial fluctuations
on earlier tasks, with large performance drops sometimes
followed by partial recovery. These effects are particularly pronounced on FOMC and MeetingBank. In contrast, \method{} generally preserves task scores more consistently and exhibits smaller deviations from the task-completion baseline. For example, on Qwen3.5-9B, LoRA initially achieves a higher MeetingBank score than \method{}, but its score falls to 16.3 after learning all eight tasks, whereas \method{} retains 57.9.
The corresponding performance drops are 48.1 and 3.4 points,
respectively. The absolute-score matrices and forgetting curves
show that \method{} maintains previously learned capabilities more consistently throughout sequential adaptation. This analysis complements the final performance and backward transfer results in Table~\ref{tab:sequential_trace} by revealing when degradation occurs and how it evolves as additional tasks are learned.

\end{document}